%% file: main.tex
\documentclass{siamart251104}
\input{shared}
\begin{document}
\maketitle
% REQUIRED
\begin{abstract}
The integration of Scientific Machine Learning (SciML) techniques with uncertainty
quantification (UQ) represents a rapidly evolving frontier in computational science.
This work advances Physics-Informed Neural Networks (PINNs) by incorporating
probabilistic frameworks to effectively model uncertainty in complex systems. Our
approach enhances the representation of uncertainty in forward problems by combining
generative modeling techniques with PINNs. This integration enables in a systematic
fashion uncertainty control while maintaining the predictive accuracy of the model.
We demonstrate the utility of this method through applications to random differential
equations and random partial differential equations (PDEs).
\end{abstract}
% REQUIRED
% REQUIRED

\section{Introduction}\label{Se:1}

\subsection{Problem Formulation and Approach}\label{SSe:1.1}
 In this paper, we consider random differential equations, which are differential 
 equations that depend on random parameters. Assuming that we have knowledge of the 
 probability distribution of these parameters, we aim to assess the probability 
 distribution of the solutions. This is a fundamental problem in uncertainty 
 quantification and can be regarded as the initial step in designing algorithms for 
 more intricate inverse problems where we aim to determine both the form of the 
 equation and its solutions with controlled uncertainty by leveraging the available 
 data.

 The primary tools we will employ include discrete neural network methods and 
 generative models, along with adaptations of ideas from Physics-Informed Neural 
 Networks tailored to our current problem. Machine learning and neural networks have 
 proven to be highly effective tools for approximating probability distributions in 
 $\mathbb{R}^k$, even for large dimensions $k$. However, the solutions of random 
 differential equations can be viewed as probability distributions on 
 infinite-dimensional spaces (typically Hilbert or Sobolev spaces), making their 
 formulation and approximation particularly intricate. This is one of the key issues 
 we aim to address in the present work.

The problems we consider have the form
\begin{equation}\label{eq:RPDE:abstarct}
\begin{split}
  &\mathscr {A}_\xi u(\cdot, \xi)=\mathscr {F}_\xi ,
\end{split}
\end{equation}
where $\mathscr {A}_\xi$ is a $\xi-$parametrised differential operator (incorporating 
possible initial and boundary conditions), and $\mathscr {F}_\xi$ is the source. For 
each fixed $\xi$, $u(\cdot;\xi)$ denotes the solution of the differential equation, 
which we interpret as a mapping $U:\Xi\to \HH$ with $U(\xi)\coloneqq u(\cdot,\xi)$. The space 
$\HH$ where $u$ takes values for each $\xi$ is an (infinite dimensional) Hilbert or 
Sobolev space. We assume that the random parameter $\xi\in \Xi$ is distributed 
according to a known Borel probability measure $\gamma$ on $\Xi$. Our aim is to 
approximate the random variable $U$ and its law $\nu _u\in \mathcal P (\HH )\, ;$ 
here $\mathcal P (\X )$ denotes the set of Borel probability measures on $\X$. 

\subsubsection*{Motivation and Neural Measures}

One of the key successes of neural network methods is their ability to effectively 
approximate probability distributions. In fact, several important problems of interest 
rely on generating a probability distribution when the available data represent 
specific instances of the unknown distribution. Typically, if the data is represented 
by an empirical measure $\nu_{\, \text{data}} \in \mathcal{P}(\mathbb{R}^k)$, we would 
like to generate an approximation of $\nu_{\, \text{true}} \in \mathcal{P}(\mathbb{R}^k)$ 
by optimising
\begin{equation}\label{mm_nn:abstract}
	  \min_{ \mu  \in  V  _{\Theta } } d_{\mathscr {M}}\lr{\mu ,\nu _{\, \text{data}} },
\end{equation}
where $V_{\Theta}$ are appropriate discrete subsets of $\mathcal{P}(\mathbb{R}^k)$ 
determined by a parameter set $\Theta$ which is connected in an appropriate manner 
with neural network sets of given architecture and $d_{\mathscr {M}}\lr{\cdot ,\cdot }$ 
a specified distance (or divergence) on $\mathcal{P}(\mathbb{R}^k)$. We then hope that
\begin{equation}\label{mm_nn:abstract:2}
\mu_{\theta ^*} \cong \nu _{\, \text{true}}, \qquad \text {where } \mu_{\theta ^*} 
\in \argmin _ {\mu \in V _{\Theta } } d_{\mathscr {M}}\lr{\mu ,\nu _{\, \text{data}} } .
\end{equation}

Representative works utilising the aforementioned generative model framework for a 
variety of applications include 
\cite{adler2018banach,
  goodfellow2014generative,
  arjovsky2017wassersteingan,
  zhang2018wassersteinganmodeltotal,
  creswell2018generative,
  jospin2022hands},
and the references in \cite{Katsoulakis_z2023meanfieldlab}. For a recent modelling 
perspective that interprets various generative models as mean field games, see 
\cite{Katsoulakis_z2023meanfieldlab}.
Recent advances connect random PDEs with foundational models \cite{Katsoulakis2025}.
 
The above framework relies on various alternative choices, each one yielding 
fundamentally distinct algorithms at the end. We would like to explore the potential 
extension of this approach to approximate solutions of \cref{eq:RPDE:abstarct}. To 
accomplish this we must first systematically construct discrete spaces $V_\Theta$ 
such that $V_\Theta \subset \mathcal{P}(\HH).$ We adopt the following alternative 
approaches:
\begin{itemize}
  \item fully NN based $V_\Theta,$ detailed in Section 3.1
  \item Polynomial Chaos Expansion (PCE)-NN $V_\Theta,$ detailed in Section 3.2
  \item Galerkin-NN $V_\Theta, $ detailed in Section 3.3.
\end{itemize}
Furthermore, we show that the neural measure spaces are expressive enough to approximate
appropriate subspaces of $\mathcal{P}(\HH)$.

\subsubsection*{Incorporating PINNs}

Once appropriate $V_\Theta, $ $V_\Theta \subset \mathcal{P}(\HH),$ are in place, one may
want to incorporate the key idea of Physics Informed Neural Networks and to combine it
with the Neural Measure spaces. In fact, one needs to define
\begin{equation}
	 \mathscr{A} \odot \mu, \qquad \mu \in V_\Theta \subset \mathcal{P}(\HH),
\end{equation}
i.e. the probability distribution on $\mathcal{P}(\HH ^{\text {imag} }),$ which is
obtained by applying $\mathscr {A}_\xi $ to $\mu$ in an appropriate sense. Here
$\mathscr {A}_\xi u(\cdot, \xi) , \mathscr {F}_\xi \in \HH ^{\text {imag} } $ for each
fixed $\xi.$ This definition is made precise in Sections 2.3 and 2.4.

We then consider the abstract problem,
\begin{equation}\label{mm_nn:abstract:pinn}
	  \min  _ {\mu  \in  V  _{\Theta } } d_{ \tilde {\mathscr {M}} }\lr{ \mathscr{A} \odot \mu
	  ,\nu _{\,  \mathscr {F} } } ,
\end{equation}
where $\nu _{\,  \mathscr {F} }\in \mathcal{P}(\HH ^{\text {imag} })$ is the distribution
corresponding to the source 
$\mathscr {F}_\xi,$ and $d_{\tilde {\mathscr {M} } }\lr{\cdot,\cdot }$ 
a specified distance (or divergence) on $\mathcal{P}(\HH ^{\text {imag} })$.

\subsubsection*{Formulation of the  methods and results}
The plan outlined above is implemented systematically in the following sections. We
adopt a concrete yet somewhat abstract approach for two main reasons:

(a) \emph{the formulation of discrete neural measure spaces has the potential to be
valuable in various other applications of uncertainty quantification (UQ) for partial
differential equations (PDEs)}, and \\

(b) \emph{the integration of the physics-informed neural networks (PINN) framework
into the final algorithms can be carried out in a general manner, independent of the
specific choice of measure metrics.} \\

\noindent We then show that employing Wasserstein distances offers particular
advantages in the implementation of the final algorithms.

This work is structured to systematically develop and analyse a generative modelling
approach for Uncertainty Quantification using Physics-Informed Neural Networks
(PINNs). In \Cref{Se:2}, we introduce the underlying problem, and the
corresponding probabilistic framework essential to our study. Next, we introduce new
notational conventions and the variational formulation used. The variational principle
forms the backbone of our approach, guiding the optimisation process for learning the
solution distribution of the random differential equation. \Cref{Se:3} presents
the model construction in detail. We define a fundamental representation of Neural
Measures tailored to the problem’s probabilistic nature and explore three distinct
architectural frameworks, namely fully discrete randomised neural networks, Polynomial
Chaos Expansion Neural Networks (PCE-NN), Galerkin-based neural networks. In 
\Cref{Se:4}, we shift focus to the variational distances used during training, with
particular emphasis on the Wasserstein distance and its interpretation in terms of
Dirac measures, which naturally relate it to $L^p$-type loss functionals. We also
discuss alternative metrics and divergences that can be integrated into the framework.
\Cref{Se:5} presents numerical experiments that validate the proposed framework
across three representative problems: a bistable ODE, a linear diffusion PDE, and a
reaction-diffusion PDE. In each case, the problem parameters are treated as random
variables, and the aim is to learn the distribution of the solution. The results
demonstrate that the generative PINN models can accurately capture both the mean and
higher-order statistical properties of the solutions. Comparisons with reference
solvers show very good qualitative and quantitative agreement, though challenges arise
in representing sharply peaked distributions and boundary behaviours. Different
architectures, including standard PINNs and PINN-PCE variants, are assessed in terms
of convergence, error behaviour, and their effectiveness in representing uncertainty.
\Cref{Se:6} presents universal approximation theorems showing that the neural measure 
spaces, suggested herein, are expressive enough.
Therein, we also provide the formal definitions and foundational elements used throughout 
the work.

\subsubsection*{\it Previous works}
Physics-Informed Neural Networks (PINNs) are a class of neural network-based methods
designed to approximate solutions of partial differential equations (PDEs), where the
loss function incorporates the residual of the PDE \cite{Karniadakis:pinn:orig:2019}.
Similar approaches have been explored in 
\cite{Lagaris_1998,
  Berg_2018,
  Raissi_2018,
  SSpiliopoulos:2018,
  makr_pryer2024deepUz}.

To the best of our knowledge, the systematic framework presented here is novel.
However, a number of related studies share certain aspects with our approach. A review
of neural network approaches to uncertainty quantification can be found in
\cite{Psaros:2023}. A method which uses the PINN framework for random differential
equations is presented as well. It is assumed the approximate solution is distributed
with a normal distribution and the method provides ways to estimate its mean and
variance.
Inferencing stochastic solution and parameters of a Random PDE, by combining Polynomial
Chaos Expansions, \cite{PCE_karniadakis}, and PCA we considered in
\cite{ZHANG2019108850,
  zhang2019learningmodalspacesolving%
  },
other related works include \cite{ping2024uncertainty}. Physics Informed Generative
Adversarial Networks were considered in \cite{gao2023pi} and \cite{yang2020physics}.
See also
\cite{meng2022learning,
  MAL-056,
  unlu2020variational%
},
for UQ related approaches. Neural Operators to address certain issues involving 
stochasticity in approximating PDEs in infinite dimensional setting include 
\cite{lu2019deeponet,
  Stuart_NOper_JMLR:24,
  Mishra_Conv_operNEURIPS2023,
  kovachki2024operator,
  li2020fourier,
  de2022cost,
  magnani2022approximate%
},
approaches were proposed in 
\cite{lu2021learning,
  Yang_2021,
  Liu_2023,
  figueres2025pinndomaindecomposition%
},
their new role and importance in
application to sciences was discussed in \cite{koumouts_UQNN_review}.

\section{The problem and its probabilistic framework}\label{Se:2}
 
\subsection{Problem and Assumptions}\label{sec:Prob}

The objective is to approximate the probability distribution associated with the 
solution of the following Random (PDE):
\begin{equation}\label{eq:RPDE:v2}
\begin{split}
&A_\xi u(x,t;\xi)=f(x,t;\xi) \quad \forall \lr{x,t}\in D\times (0,T]\\
&B_\xi u(x,t;\xi)=g(x,t;\xi) \quad \forall \lr{x,t}\in \Gamma
\end{split}
\end{equation}
where $u(x,t;\xi)$ denotes the solution of the PDE, which we interpret as the 
mapping $U:\Xi\to \HH$ with $U(\xi)\coloneqq u(\cdot,\xi)$. Here, $A_\xi:\HH\to\tilde\HH$ 
is a $\xi-$parametrised differential operator, and $B_\xi:\HH\to\overline\HH$ is 
the associated boundary operator. The random parameter $\xi\in \Xi$ is distributed 
according to a Borel probability measure $\gamma$ on $\Xi$. 

Analysing the behaviour of solutions is beyond the scope of this work. Our 
objective is to propose numerical algorithms for approximating the distribution 
associated with the solution of the problems at hand. Therefore, we assume that 
the solution to \cref{eq:RPDE:v2} exists and is sufficiently regular for our 
purposes. In particular, we assume throughout that 

\begin{itemize}
\item
 The operators $A_\xi$ and $B_\xi$ are well-defined for 
$\gamma-$almost every $\xi$,
and the solution of  \cref{eq:RPDE:v2} belongs in  
$L^2(\gamma; \, \HH)$.
\item 
For   $q\in L^2(\gamma;\HH)$, we assume that the following   maps:
\begin{equation}\label{hypothesis:3}
\begin{split}
& \mathcal{A} Q(\xi) \coloneqq  A_\xi q(\xi),\quad \mathcal{A} Q(\xi)
  \coloneqq  B_\xi q(\xi)\qquad\text{ are } L^2-\text{integrable, i.e., }\\[3pt]
& \mathcal{A} Q \in L^2(\gamma;\tilde{\HH})\quad\text{ and }\quad
\mathcal{B} Q\in L^2(\gamma;\overline{\HH}).
\end{split}
\end{equation}
\end{itemize}

Where the space   $ L^2(\gamma;\HH)$ is defined in the sense of Bochner integration, 
see for example Section 7.2.2 of \cite{Dashti_Stuart2017} for details. In particular this implies that 
\begin{equation*}
  \int \| U (\xi ) \| _{\HH}^2 \, \gamma (\ud \xi ) < \infty . 
\end{equation*}
and  we denote by $U$ the element of  $L^2(\gamma; \, \HH)$, corresponding to the 
solution $u.$
Using the definition of pushforward, we define its probability measure 
$\nu_u\in \mathcal P (\HH )$ by 
\begin{equation*}
  \nu_u(A)= \pushforward{U}{\gamma}(A) = \gamma\circ U^{-1}(A),\qquad \forall A\in 
  \Bor{\HH}.
\end{equation*}
Since we assume $U\in L^2(\gamma; \, \HH)$, this probability has a finite second moment.

In the next Sections, we introduce a variational method based on a parametrised family 
of Neural Models to construct an approximation of the target probability distribution 
$\nu_u$.

\subsection{Probabilistic framework}

We review some standard concepts from probability theory which will be useful in the 
sequel. We begin by assuming an abstract probability space $\lr{\Omega,\FF,\p}$, a 
measurable space over $\HH$, eg. $\lr{\HH,\Bor{\HH}}$, and a random element 
$\tilde{q}:\Omega\to\HH$, i.e., $\FF/\Bor{\HH}-$measurable function. The law of 
$\tilde{q}$ is the pushforward measure of $\p$ under $\tilde{q}$, defined as:
\begin{equation*}
  \pushforward{\tilde{q}}{\p}(A) = \p\lr{\tilde{q} \in A} = \p\circ \tilde{q}^{-1}(A),
  \qquad A\in\Bor{\HH},
\end{equation*}
where $\tilde{q}^{-1}$ denotes the preimage of the set $A$. It is straightforward to 
verify that this defines a valid probability measure.

An important assumption in our setting is that we only consider solutions such that 
$u\in L^2\lr{\gamma; \, \HH}$, which ensures two key properties:
\begin{itemize}
\item The mapping $U(\xi)=u(\cdot;\xi)$  is measurable with respect to $\xi$,
\item $U(\xi)$ has a finite second moment with respect to $\gamma$. 
\end{itemize}

Let $\xi:\Omega\to\Xi$ be a random element ( i.e.,  $\FF/\Bor{\Xi}-$measurable function) 
with law $\gamma$. We are interested in random elements $\tilde{q}$ expressible as 
compositions of a function $q\in  L^2\lr{\gamma; \, \HH}$ and $\xi$, i.e., 
\begin{equation}\label{stochastic:representation:gamma:v2}
\tilde{q}(\omega) = q\lr{\xi(\omega)},\qquad\p-\text{as } \omega.
\end{equation}
It is also straightforward to verify that $\tilde{q}$ has a finite second moment with 
respect to $\p$, and its law satisfies:
\begin{equation*}
  \pushforward{\tilde{q}}{\p}(A)= \pushforward{q}{\gamma}(A),\qquad\forall A \in 
  \Bor{\HH}.
\end{equation*}
Hence setting $\nu_q \coloneqq  \pushforward{q}{\gamma}$ we have
\begin{equation*}
  \nu_q = \pushforward{q}{\gamma} \in \mathcal {P} (\HH),
\end{equation*}
where $\PPP(\HH)$ denotes the set of all Borel probability measures on $\HH$, i.e.,  
measures defined on $\lr{\HH,\Bor{\HH}}$. 

Since our goal is to approximate probability measures, we introduce the following 
collections:
\begin{itemize}
\item For $p\geq 1$, define the sub-collection $\PPP_p\lr{\HH}$ consisting of all 
measures with finite $p-$th moment,
\begin{equation*}
  \PPP_p\lr{\HH}\coloneqq \lrr{%
  \begin{split}
  \mu\in\PPP\lr{\HH}:\,\int_\HH d_{\HH}(h,\overline{h})&^p\mu(\ud h)<\infty \\
   &\text{for some (and hence any) }\overline{h}\in\tilde\HH
   \end{split}
   },
\end{equation*}
where $\lr{\HH,d_\HH}$ is the metric space generating the Borel 
$\sigma-$algebra $\Bor{\HH}$.
\item 
Given a reference measure $\gamma$ on $\Bor{\Xi}$, define sub-collection of $\PPP_2\lr{\HH}$ 
\begin{equation*}
  \PPP_p^\gamma\lr{\HH} = \lrr{\mu\in\PPP(\HH):\,\mu=\pushforward{X}{\gamma}\,
  \text{ for some }X\in L^p\lr{\gamma;\HH} }.
\end{equation*}
\end{itemize}
Over those collections, one  may consider alternative metrics for distances or divergences. 
In this work, we shall use  the Wasserstein distance $W_p$, 
\cite{AGS:2005, Villani_OT_book_2009} and the corresponding  $\lr{\PPP_p\lr{\HH},W_p}$   
metric space. The $p-$Wasserstein distance is given by
\begin{equation}\label{Wasser:distance}
W_p(\mu,\nu)\coloneqq \inf_{\pi\in\Pi(\mu,\nu)} \lr{\int_{\HH\times\HH} 
\,d_\HH(x,y)^p \,\,\pi(\ud x,\ud y)}^{1/p} ,
\end{equation}
where $\Pi(\mu,\nu)$ is the set of all plans (joint distributions) with marginals $\mu$ 
and $\nu$, \cite{AGS:2005}. Furthermore, we shall use the notation, 
\begin{equation}\label{distance:p:g}
\begin{split}
d_{p,\gamma}(\pushforward{X}{\gamma}, \pushforward{Y}{\gamma})^p :&=\int_{\HH\times \HH} 
d_{\HH}(x,y)^p\,\pushforward{\lr{X,Y}}{\gamma}(\ud x,\ud y)\\ 
&= \int_{\Xi} \norm{X(\xi) -Y(\xi)}{\HH}^p\,\gamma(\ud \xi).
\end{split}
\end{equation}
The distance  $d_{p,\gamma}$ induces a metric on  $L^p(\gamma;\HH).$

\subsection{Transformed Law due to Parameterised Operators}\label{def:odot}

We now introduce a notation to represent the law of expressions such as $A_\xi q(\xi)$ 
and $B_\xi q(\xi)$. In the most general setting, one might define stochastic processes 
$\tilde{A}: \Xi\times\HH\to\tilde\HH$ and $\tilde{B}:\Xi\times\HH\to\overline\HH$, with 
appropriate measurability conditions.
However, in \Cref{sec:Prob}, we simplify the setting by assuming hypothesis 
stated in \cref{hypothesis:3}, ensuring measurability. We then define the operation 
$\odot$ as follows:
\begin{equation*}
  A \odot \lr{\pushforward{q}{\gamma}}(C) \coloneqq  \pushforward{A_\xi q(\xi)}{\p}(C) ,
\end{equation*}
and similarly for the parameterised operator $B_\xi$. 
For instance we may rewrite the random PDE as equality of probability measures as follows,
\begin{equation*}
  A \odot \nu_u(C)
= \nu_f(C)  = \pushforward{f(\cdot;\xi)}{\p}(C),\qquad \forall C\in\Bor{\tilde\HH}.
\end{equation*}
It will be useful to summarise these definitions as follows. 
We define and study metric spaces of Borel probability measures, such as  
$\PPP_2^\gamma(\HH),\,\PPP_2^\gamma(\tilde\HH)$ and $\PPP_2^\gamma(\overline\HH)$, 
and decompose the law of the solution ${u}$ into a measurable function $U(\xi)$ and 
a probability measure $\gamma$, according to the diagram:
\begin{equation*}
  \p\quad\xrightarrow{\xi}\quad\gamma\quad
  \begin{cases}
    \xrightarrow{q}\quad\pushforward{q}{\gamma}\\
    \xrightarrow{A\odot\,}\quad A\odot\,\pushforward{q}{\gamma} \\
    \xrightarrow{B\odot\,}\quad B\odot\,\pushforward{q}{\gamma},
  \end{cases}
\end{equation*}
with 
\begin{gather*}
\p \in \PPP(\Omega),\qquad \gamma\in\PPP(\Xi),\quad\pushforward{q}{\gamma}\in 
 \PPP_2^\gamma(\HH)\\[2pt]
A\odot\,\pushforward{q}{\gamma} \in \PPP_2^\gamma(\tilde\HH)\quad\text{and}\quad 
A\odot\,\pushforward{q}{\gamma} \in \PPP_2^\gamma(\overline\HH) .
\end{gather*}

\subsection{Variational Approximation of the Target Law}

To approximate the target probability measure $\nu_u$, we adopt a variational approach. 
Specifically, we minimise a discrepancy functional over $\VV _\Theta =\VV_\NN^*$ that 
quantifies the difference between transformed distributions of the model and the target, 
induced by the operators $A_\xi$ and $B_\xi$. Alternative specific choices for the neural 
measure spaces $\VV _\Theta =\VV_\NN^*$ are provided in the next section. The discrepancy 
we minimise is given by:
\begin{equation*}
  d_{\PPP_2^\gamma(\tilde\HH)}\lr{A \odot\nu_u,A \odot\mu_\theta} + 
  d_{\PPP_2^\gamma(\overline\HH)}\lr{B \odot\nu_u,B \odot\mu_\theta} ,
\end{equation*}
where $d_{\PPP_2^\gamma(\tilde\HH)}$ and $d_{\PPP_2^\gamma(\overline\HH)}$ are appropriate 
metrics defined over the space of probability measures $\PPP_2^\gamma\lr{\tilde\HH}$ and 
$\PPP_2^\gamma\lr{\overline\HH}$, respectively. The variational problem then becomes:
\begin{equation} \label{var:problemm}
  \mu_{\theta^*} \in \mathop{\mathrm{arg\,min}}_{\mu_\theta \in \mathcal{V}_{\mathcal{N}}^*}
  d_{\mathcal{P}_2^\gamma(\tilde{\mathcal{H}})} (A \odot \nu_u, A \odot \mu_\theta) + 
  d_{\mathcal{P}_2^\gamma(\overline{\mathcal{H}})} (B \odot \nu_u, B \odot \mu_\theta).
\end{equation}
By varying the choice of metrics, neural architectures $\VV_\NN$ and the structure of 
$X_\theta$, different approximation strategies can be implemented. Nonetheless, the 
ultimate objective remains the same: to identify an optimal Neural Measure $\mu_{\theta^*}$ 
that is close enough to the target distribution $\nu_u$.

\section{Discrete Neural Measure Spaces}\label{Se:3}

To motivate the definition of neural measures on $\mathcal {P} (\HH )$ we consider a family
of discrete stochastic mappings
\begin{equation*}
  \tilde{X}_\theta:\Omega \to \HH,
\end{equation*}
where each map $\tilde{X}_\theta$ is \emph{associated} with a neural network function
parameterised by $\theta\in\Theta$, where $\Theta$ indexes a class of neural networks
$\VV_\NN$. Our building assumption is that $\tilde{X}_\theta$ will be based on standard
discrete neural spaces $\VV_\NN$ designed to approximate functions. The alternative
designs we consider are based on neural network functions corresponding to (i) both
deterministic ($x, t$) and stochastic variables ($\xi$), (ii) only deterministic
variables and (iii) only stochastic variables. Thus, each model will be characterised
by a function $g_\theta\in\VV_\NN$, and our goal is to approximate the distribution
associated to the solution $u$ by solving an appropriate optimisation problem over
$\Theta$.

To be more precise, let $\VV_\NN$ be a chosen class of neural networks as above, and let
$\gamma$ be the reference probability measure defined on a latent space $\Xi$. We define
first a corresponding family of functions $X_\theta = X_\theta (x, t, \xi),$
\begin{equation*}
  X_\theta:\Xi \to \HH ,
\end{equation*}
where each model $X_\theta$ is implicitly determined by a function $g_\theta\in\VV_\NN,$
i.e., $X_\theta =X ( g_\theta).$ These functions encode the neural architecture and
the parameterisation of the model, and will be detailed in the following Sections
3.1-3.

To introduce stochasticity, we consider a random variable $\xi:\Omega\to \Xi$, distributed
according to $\gamma$. Composing this with the deterministic map $X_\theta$ yields the
stochastic model:
\begin{equation*}
  \tilde{X}_\theta(\omega) = X_\theta \circ \xi(\omega) .
\end{equation*}
This composition defines a random element in $\HH$ allowing us to generate samples from
the associated distribution. The pushforward of $\gamma$ through $X_\theta$, denoted by
$\pushforward{X_\theta}{\gamma}$, defines the induced discrete Neural Measure Space on
$\mathcal {P} (\HH).$ In fact, the class of Neural Measures spaces we consider have the
form,
\begin{equation*}
  \VV_\NN^* = \lrr{ \mu_\theta = \pushforward{X_\theta}{\gamma} \text{ for some Neural
  Model }X_\theta= X ( g_\theta), \, \ \ g_\theta\in\VV_\NN \, }.
\end{equation*}
Clearly, $\VV_\NN^* \subset \mathcal {P} (\HH)$ and in fact $\VV_\NN^* \subset
\PPP_2^\gamma\lr{ \HH\,} .$ Here, $\VV_\NN$ is fixed a priori, and the notation
$\VV_\NN^*$ reflects the fact that this space is associated with a discrete standard
neural network space $\VV_\NN.$ Each model in $\VV_\NN^*$ corresponds to a unique
pushforward measure derived from a neural map and the reference measure. Notice finally
that although $\xi\sim \gamma$ is sampled from a known reference distribution, its
explicit form $\xi\lr{\omega}$ need not to be specified.

\subsection{Fully Network Based Neural Measure Spaces}\label{sec:arch:gamma}

Next, we consider $X_\theta$, defined in relation to neural network functions 
$g_\theta\in\VV_\NN$, corresponding to both deterministic ($x, t$) and stochastic 
variables ($\xi$). The resulting neural models solely rely on a chosen class of neural 
networks $\VV_\NN$ without any further approximation step.

By fixing the number of layers, the size of each hidden layer, and the activation 
functions, we define the following class 
\begin{equation*}
  \mathcal{V}_{\mathcal{N}} \coloneqq \left\{ 
  \begin{aligned}
      g_\theta : \mathbb{R}^{d_{\text{in}}} \times \mathbb{R}^{d_{\text{ran}}} &\to \mathbb{R}^s \\
      (x,t;\xi) &\mapsto (C_L \circ \sigma_{L-1} \circ \dots \circ \sigma_1 \circ C_1)
      (x,t;\xi)
  \end{aligned}
  \;\; : \;\; \theta \in \Theta \right\} .
\end{equation*}

In the above procedure, any such map $g_\theta$ is defined by the intermediate layers 
$C_k$, which are affine maps of the form
\begin{equation} \label{C_k}
  C_k y = W_k y + b_k, \quad \text{with} \quad W_k \in \mathbb{R}^{d_{k+1} \times d_k}, 
  \;\; b_k \in \mathbb{R}^{d_{k+1}},
\end{equation}
where the dimensions $d_k$ may vary with each layer $k$ and the activation function 
$\sigma (y)$ denotes the vector with the same number of components as $y$, where 
$\sigma (y)_i= \sigma(y_i).$ The index $\theta\in \Theta $ represents collectively 
all the parameters of the network.

Then, we define the neural model as:
\begin{equation*}
  X_\theta(\xi) = g_\theta\lr{\cdot;\xi}, \text{ for some }
  g_\theta\in \VV_\NN.
\end{equation*}
We may additionally require that for every $g_\theta\in \VV_\NN$
\begin{itemize}
\item $g_\theta(\cdot,\xi)\in \HH$, for every $\xi$,
\item and 
$\int_\Xi\norm{g_\theta(\cdot;\xi)}{\HH}^2\gamma(\ud \xi) <\infty $.
\end{itemize}
These ensure that $X_\theta \in L^2(\gamma;\HH)$. The associated variational method is 
optimised based on the following class of Neural Measures:
\begin{equation*}
  \VV_\NN^* = \lrr{\pushforward{X_\theta}{\gamma}\in\PPP_p^\gamma\lr{\,\,\HH\,\,}: 
  \text{ where }X_\theta(\xi)=g_\theta(\cdot,\xi) \text{ for some }g_\theta\in \VV_\NN}.
\end{equation*}
This formulation is simple and general--it captures a wide range of model classes 
without assuming any structure other than that provided by the neural network 
architecture itself. The construction presented is typical, and several alternative 
architectures are feasible; it is presented in this specific form for completeness of 
exposition.

\subsection{PCE-NN  Based Neural Measure Spaces}\label{sec:truncated:basis}

In this section, we introduce a class of neural models designed to approximate functions 
$U\in L^2(\gamma;\HH)$, where the computational complexity of learning high-dimensional 
mappings $\lr{x,t;\xi}\mapsto \R^{s}$ is reduced through truncated spectral expansion. 
The key idea is to exploit the linear structure of the Hilbert space $L^2(\gamma;\HH)$ 
and separate the stochastic and deterministic components of the function.

To this end, consider a Polynomial Chaos Expansion (PCE)-type basis, 
\cite{PCE_karniadakis}, $\lrr{\phi_n}_{n\in\N}\subset L^2(\gamma)$ satisfying:
\begin{equation}\label{PCE:base}
\forall q\in L^2(\gamma;\HH),\quad\exists\,!\,\lrr{a_n}_{n\in\N}\subset\HH: \quad 
q(x,t;\xi) = \sum_{n\in\N} a_n(x,t) \phi_n(\xi)\vspace{-0.3cm} ,
\end{equation}
we represent the neural model as a truncated expansion:
\begin{equation*}
  X_\theta(\xi) = \sum_{n=1}^{K } g_\theta^{(n)}\,\phi_n(\xi) ,
\end{equation*}
where each coefficient function $ g_\theta^{(n)}\in\HH$ is a component of the output 
layer of some neural network $g_\theta$ within $\VV_\NN$, i.e.
\begin{equation}\label{PCE:VN}
  \VV_\NN \coloneqq \lrr{g_\theta: \R^{d_{\text{in}}}\to\R^{d_{\text{out}}K}: \ g_\theta =g_\theta(x,t)
  \text{ for   }\theta\in\Theta },
\end{equation}
and the structure of the neural network functions $g_\theta(x,t)$ is as before, but now 
affecting only $x, t$ variables. Then, the corresponding form of neural measures is: 
\begin{equation}\label{PCE:VNstar}
  \VV_\NN^* = \lrr{\pushforward{X_\theta}{\gamma}\in\PPP_2^\gamma\lr{\,\,\HH\,\,}: 
  \text{ where }X_\theta(\xi)=\sum_{n=1}^{K } g_\theta^{(n)}\,\phi_n(\xi) \text{ for some }
  g_\theta\in \VV_\NN}.
\end{equation}
This construction decouples the stochastic and deterministic components, enabling: 
simpler and lower-dimensional neural architectures, and possibly structured learning 
problems.

\subsection{Galerkin-NN Based Neural Measure Spaces}\label{sec:GalerkinNN}

Galerkin-NNs are conceptually similar to PCE-PINNs but draw inspiration from the 
Galerkin method used in numerical PDEs. In this approach, the governing equations 
are projected onto a test basis, and neural networks are formulated accordingly. 
The architecture closely mirrors that of PCE-NNs, although neural networks are 
used now for the discretisation of $\xi$ variable. For this reason, we only 
present the analogue corresponding to the variant that directly approximates the 
solution $U(\xi)$.

Consider appropriate a given finite dimensional Galerkin subspace of $\HH$, denoted 
by $\HH_h,$ let $M$ be its dimension and $\lrr{\psi_n}_{1}^M$ its basis. Then, we 
consider the elements of $L^2(\gamma;\HH)$
\begin{equation}\label{Galerkin:base}
  q(x,t;\xi) = \sum_{n=1}^M b_n(\xi) \psi_n(x,t) . 
\end{equation}
Clearly, for each fixed $\xi, $ $ q(\cdot;\xi) \in \HH_h.$ We then represent the 
neural model as  
\begin{equation*}
  X_\theta(\xi) = \sum_{n=1}^{M } g_\theta^{(n)}\lr{\xi}\,\psi_n ,
\end{equation*}
where each coefficient function $g_\theta^{(n)}\in L^2(\gamma)$ is a component of 
the output layer of some neural network $g_\theta$ within $\VV_\NN$, i.e.,
\begin{equation*}
  \VV_\NN \coloneqq \lrr{g_\theta: \R^{d_{\text{ran}}}\to\R^{M}:
   g_\theta= g_\theta(\xi) \text{ for 
  }\theta\in\Theta },
\end{equation*}
and the structure of the neural network functions $g_\theta(\xi)$ is as before, but 
now affecting only the $\xi$ variable. Then, the corresponding form of neural 
measures is: 
\begin{equation*}
  \VV_\NN^* = \lrr{\pushforward{X_\theta}{\gamma}\in\PPP_2^\gamma\lr{\,\,\HH\,\,}: 
  \text{ where }X_\theta(\xi)=\sum_{n=1}^{M} g_\theta^{(n)}\lr{\xi}\,\psi_n \text{ for 
  some }g_\theta\in \VV_\NN}.
\end{equation*}
Although the above formulation is quite flexible and compatible with more standard 
methods for the numerical solution of PDEs, its actual implementation depends on the 
choice of the spaces $\HH _h. $ It is possible using the Discontinuous Galerkin 
framework, to consider spaces that are not subspaces of $\HH$. This approach is 
more involved and is beyond the scope of this work.

\section{Distances and their evaluation} \label{Se:4}

In statistical learning and probabilistic modeling, measuring the "distance" between 
probability measures is a fundamental task. Two broad families of such distances are 
commonly used: divergences and metrics, both of which serve to quantify dissimilarity 
over a suitable space of probability measures, such as $\PPP(\HH)$. Depending on the 
nature of the problem, several alternative choices can be made, resulting in 
different loss functions. Our framework is flexible in this regard. We will focus 
next on Wasserstein distances, as in this case, the loss function takes a particular 
simple form.

\subsection{Loss based on Wasserstein distance}

Within the framework the present paper it is particularly convenient to consider 
Wasserstein distances. A key observation is that when one of the measures to be used 
in the loss is atomic, i.e., $\delta_g,$ then the set of plans is reduced to 
$\mu \times \delta _g,$ i.e., $\Pi (\mu, \delta_g) = \{ \mu \times \delta _g \}$,
see (5.2.12) in \cite{AGS:2005}.    
Assuming $g\in\HH$ and $\delta_g \in \PPP_p(\HH)$, we then have
\begin{equation}
	\begin{split}
		 W_p(\mu,\delta_g)^p = & \inf_{\pi\in\Pi(\mu,\delta_g)} \int_{\HH}\int_{\HH} 
		 \norm{x-y}{\HH}^p \pi(\ud x, \ud y) \\
     =& \int_{\HH}\int_{\HH} \norm{x-y}{\HH}^p 
		 \mu(\ud x)\delta_g( \ud y)\\
		  =& \int_{\HH} \norm{x-g}{\HH}^p \mu(\ud x) = \E_{X\sim \mu} \norm{X-g}{\HH}^p .
	\end{split}
\end{equation}
We distinguish two cases. 
In case where $\mathcal{A} U(\xi)$ and $\mathcal{B} U(\xi)$ are constant with respect to 
$\xi$, we can easily check exists $f\in\tilde{\HH}$ and $g\in\overline\HH$ such 
that 
\begin{equation*}
  A \odot\nu_u = \delta_{f},\qquad B \odot\nu_u =\delta_g ,
\end{equation*}
and the suggested variational method can be written as follows:
\begin{equation*}
  \mu_{\theta^*} \in \argmin_{\mu_\theta \in \mathcal{V}_{\mathcal{N}}^*} 
  W_p(A \odot \mu_\theta, \delta_f) + W_p(B \odot \mu_\theta, \delta_g)
\end{equation*}

In the general case, it is still possible to use this property if we define residual 
operators
\begin{equation*}
  \tilde{A}_\xi X_\theta(\xi) \coloneqq  \mathcal{A} X_\theta(\xi) - \mathcal{A} U(\xi),
  \quad\text{ and }\quad
  \tilde{B}_\xi X_\theta(\xi) \coloneqq  \mathcal{B} X_\theta(\xi) - \mathcal{B} U(\xi) ,
\end{equation*}
and the suggested variational method can be written as follows:
\begin{equation*}
  \mu_{\theta^*} \in \argmin_{\mu_\theta\in\VV_{\NN}^*}\,\, W_p(\tilde{A} 
  \odot\mu_\theta,\delta_0)+ W_p(\tilde{B} \odot\mu_\theta,\delta_0) .
\end{equation*}

This is particularly helpful, since the computation of the Wasserstein distances in 
both cases is reduced to ($p=2$) the evaluation of integrals of functions, 
\begin{equation*}
\begin{split}
  d_{2,\gamma}\lr{A\odot\nu_u,\,A\odot\mu_\theta}^2 +& d_{2,\gamma}\lr{B\odot\nu_u,\,
  B\odot\mu_\theta}^2 \\
  &= \norm{\mathcal{A} U - \mathcal{A} X_\theta }{L^2(\gamma;\tilde\HH)}^2+
      \norm{\mathcal{B} U - \mathcal{B} X_\theta }{L^2(\gamma;\overline\HH)}^2.
\end{split}
\end{equation*}
In fact, recall that both $\nu_u,\,\mu_\theta \in \PPP_2^\gamma(\HH)$. Additionally, 
by hypothesis \cref{hypothesis:3}, it follows that $A\odot\nu_u,\,A\odot\mu_\theta 
\in \PPP_2^\gamma(\tilde\HH)$ and $B\odot\nu_u,\,B\odot\mu_\theta \in 
\PPP_2^\gamma(\overline\HH)$.

\subsection{Other metrics and divergences}

It is possible, however, to utilise alternative divergences or metrics to define the 
loss functional. We briefly review some potential choices below, see, eg., 
\cite{Katsoul_pantazis2022,
  renyi1961measures,
  muller1997integral,
  sriperumbudur2009integral,
  birrell2022f,
  Villani_OT_book_2009,
  sen2011introduction}.

\subsubsection{Divergences and Metrics: Definitions and Relationships}

Divergences generalise metrics by dropping symmetry and the triangle inequality, while
retaining non-negativity and vanishing only when the inputs are equal. A key family is
the f-divergences: given probability measures $\mu\ll\nu$ and a convex function
$f:\R_{+}\to\R\cup\lrr{+\infty}$ with $f(1)=0$, the $f-$divergence is defined as
\begin{equation*}
  D_f(\mu|| \nu ) = \int f\lr{\frac{\ud \mu}{\ud \nu}} \ud \nu .
\end{equation*}
This includes the
Kullback–Leibler (KL) divergence, Jensen–Shannon divergence, and others. In addition to
divergences, several frameworks exist for constructing metric distances between
probability measures. These include:
\begin{itemize}
\item[(A)] 
Integral Probability Metric (IPM):

Given a class of test functions $\FF$, the IPM between two measures $\mu,\,\nu$ 
is defined as:
\begin{equation}\label{IPM:def}
d_{\FF}(\mu,\nu)\coloneqq 
\sup_{f\in\FF} \left | \E_{X\sim\mu} f(X) - \E_{Y\sim\nu} f(Y) \right | .
\end{equation}
\item[(B)] 
Optimal Transport (OT) Metrics:

In OT theory, distances are defined via
couplings. Given a cost function $c(x,y)$, the Wasserstein distance (or more generally,
$c-$Wasserstein distance) is 
\begin{equation*}
  W_c(\mu,\nu) = \inf_{\pi\in \Pi(\mu,\nu)} \int c(x,y) \ud\pi(x,y) ,
\end{equation*}
where $\Pi(\mu,\nu)$ is the set of couplings of $\mu$ and $\nu$.
\item[(C)] 
TV Distance via Densities: 

Given a probability measure
$\lambda\in\PPP(\HH)$, we define $\PPP_{\lambda}(\HH)\subset\PPP(\HH)$ to be the
collection of probability measures that are absolutely continuous with respect to
$\lambda$, i.e.,  $\mu\ll\lambda$. Let $\mu,\,\nu\in\PPP_{\lambda}(\HH)$ with Radon-Nikodym
derivatives $f_\mu,\,f_\nu$ with respect to $\lambda$. Then, TV distance between
$\mu,\,\nu$ can be expressed as:
\begin{equation*}
  d_{TV}(\mu,\nu)\coloneqq \frac{1}{2}\norm{f_\mu-f_\nu}{L^1(\lambda)}.
\end{equation*}
\end{itemize}

\subsubsection{Approximating Distances in Finite-Dimensional Spaces}

Given that our spaces, $V_\Theta = V_{\mathcal N} ^*$, have a finite number of 
parameters, we can employ techniques developed in approximating distances of 
distributions in finite dimensions to compute the actual loss function. Indicative 
such methods include, GANs and computational optimal transport algorithms, 
\cite{goodfellow2014generative, Peyre:CompOT:2019}.

For approximating Wasserstein distances, one may use Sinkhorn Algorithm which 
introduces entropic regularisation for efficient computation and other approaches 
detailed in \cite{Peyre:CompOT:2019}. These approaches exploit the geometric and 
variational structure of optimal transport theory and are particularly relevant in 
generative modelling and distributional approximation tasks.

Generative Adversarial Networks (GANs) provide a flexible variational framework for 
approximating probability distributions by minimising statistical distances, notably 
Integral Probability Metrics (IPMs) and, under certain conditions, {\small $f$-divergences}. 

Given a class of generators $g_\theta\in \VV_\NN$, a class of discriminators 
$f_\phi\in \overline\VV_\NN$, a reference distribution $\pi_0$ and target distribution 
$\p_r$ the standard GAN objective takes the form:
\begin{equation*}
  \min_{g_\theta\in \VV_\NN} \max_{f_\phi\in \overline\VV_\NN}\,\, \E_{X\sim\p_r} 
  f_\phi(X) - \E_{Z\sim\pi_0} f_\phi\circ g_\theta\circ(Z) .
\end{equation*}
In the special case where 
the class of test functions $\FF$ associated with a particular IPM (eg. as in equation 
\cref{IPM:def}) can be approximated by the class of discriminators $\overline\VV_\NN$. 
Then, this min-max formulation can be viewed as minimising IPM between $\p_r$ and 
generators distribution. Specifically, we define the class of neural measures:
\begin{equation*}
  \VV_\NN^*\coloneqq \lrr{\pushforward{g_\theta}{\pi_0}:\text{ for some }g_\theta\in\VV_\NN} .
\end{equation*}
where $\pushforward{g_\theta}{\pi_0}$ is the law of the distribution $g_\theta\circ(Z)$ 
when $Z\sim\pi_0$. The above min-max formulation provides a way to approximate the 
variational problem: 
\begin{equation*}
  \argmin_{\mu_\theta \in \VV_\NN^*} d_{\FF}(\p_r, \mu_\theta),
  \quad\text{ where }d_{\FF}\text{ is the IPM induced by the class }\FF.
\end{equation*}

\section{Numerical results}\label{Se:5}

In this section, we present numerical experiments that illustrate the performance of 
the methods introduced in Sections 2 and 3. We consider three representative 
examples: a bistable ordinary differential equation (ODE), a diffusion partial 
differential equation (PDE), and a reaction-diffusion PDE. In all cases, the problem 
parameters are treated as random variables drawn from prescribed distributions, and 
the objective is to quantify the resulting uncertainty in the solution.

All neural networks employed are fully connected multi-layer perceptrons (MLPs). The 
architectures used in the three examples consist of 5, 6, and 6 hidden layers, with 
32, 20, and 40 hidden units per layer, respectively. We use the sinusoidal-based 
activation function known as the snake activation, defined as
\begin{equation*}
  \sigma(x; a) = x + \frac{\sin^2(a x)}{a} ,
\end{equation*}
following \cite{Ziyin:2020}. This choice consistently yielded better convergence and 
expressiveness than the traditional hyperbolic tangent activation, especially for 
problems involving both periodic and non-periodic solutions. The parameter $a$ is 
specific to each layer and is optimized jointly with the network weights during 
training.

Training is performed using the L-BFGS optimizer with a learning rate of 1, a maximum
number of iterations set to 20, and a history size of 20. No batch processing is used;
the optimizer is applied to the full dataset at each step. We initialize the weights
using Xavier initialization. Training samples from the spatiotemporal domain of the
differential equation are resampled every predefined number of iterations, while
samples from the parameter space are resampled every different predefined interval.
The resampling frequency for the domain is higher than that for the parameter space.
This strategy allows the network to first focus on learning the PDE structure for a
specific parameter configuration and later generalize to new configurations as training
progresses.

All experiments were conducted using a unified codebase, which is publicly available 
at \cite{arampatzis:2024}, ensuring reproducibility and ease of extension to other 
problem settings.

\subsection{Bistable ordinary differential equation}

We consider the ordinary differential equation,
\begin{equation} \label{eq:bistable_ode}
  \frac{\mathrm{d}u}{\mathrm{d}t} = -r(u - 1) (2 - u) (u - 3), \quad t\in[0, 8],
\end{equation}
with initial condition,
\begin{equation}
  u(0) = u_0 .
\end{equation}
The parameters $u_0, r$ follow uniform distributions, $u_0 \sim \mathcal{U}(0,4)$
and $r\sim \mathcal{U}(0.8, 1.2)$. The equation has two stable ($u=1$, $u=3$) 
and one unstable ($u=2$) equilibrium solutions.

In \Cref{fig:bistable_ode_loss}, we plot the training 
and testing loss functions throughout the optimisation. 
The spatial-temporal domain is resampled every 50 iterations, 
as are the parameters, which are drawn from their predefined uniform distribution. Each 
resampling uses 100 random samples from the domain. After each resampling step, the loss 
typically spikes due to the shift in training data but continues to decrease overall as 
training progresses. The testing loss begins to rise after around 400 iterations, 
indicating that the network starts to overfit the training data.
%%%%%%%%%%%%%%%%%%
%%%%%%%%%%%%%%%%%%

\begin{figure}
    \centering
    \includegraphics[width=0.6\textwidth]{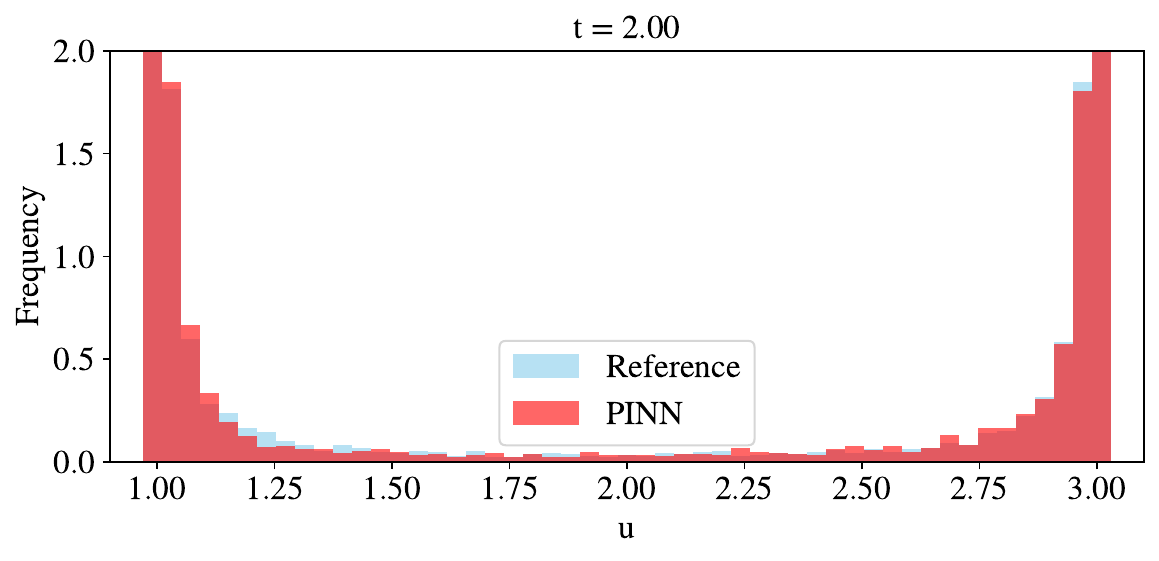}
    \caption{Histogram comparison at time $t=2$ of the solution of the bistable ODE~\cref{eq:bistable_ode} 
    using an implicit numerical solver and the PINN architecture for different parameter values.}
\label{fig:bistable_ode}
\end{figure}

In \Cref{fig:bistable_ode}, we plot histograms of the solution to the bistable ODE
\cref{eq:bistable_ode}. The reference solution is obtained using an implicit numerical
solver. We observe good qualitative agreement between the reference and the PINN-based
histograms. However, the Wasserstein distance between the two solutions increases over
time. This is expected, as the solution's distribution gradually concentrates into two
Dirac delta peaks at $u = 1$ and $u = 3$. This sharp concentration makes convergence
challenging, since the Wasserstein distance is highly sensitive to samples that have not
fully collapsed onto the Dirac masses.

\subsection{Diffusion equation}

We solve the diffusion equation
\begin{equation} \label{eq:diffusion}
  u_t - \frac{a}{k^2} u_{xx} = 0 , \quad t\in[0, 1], \quad x\in[0, \pi] ,
\end{equation}
with initial and boundary conditions,
\begin{align}
  u(0, x) &= \sin(k x), \quad x\in[0, \pi], \\
  u(t, 0) &= 0, \quad t\in[0, 1], \\
  u(t, \pi) &= e^{-a t} \sin(\pi k), \quad t\in[0, 1],
\end{align}
and exact solution,
\begin{equation}
 u(t, x)  = \mathrm{e}^{-k t} \sin (\pi x) .
\end{equation}
We assume that the parameters $a,k$ follow uniform distributions, $a\sim\mathcal{U}(1, 3)$ 
and $k\sim\mathcal{U}(1, 3)$.
%%%%%%%%%%%%%%%%%%
%%%%%%%%%%%%%%%%%%

In \Cref{fig:diffusion_loss}, we plot the training 
and testing loss functions during the training of both the PINN and PINN-PCE 
architectures for the diffusion equation \cref{eq:diffusion}. 
The spatial-temporal domain is resampled every 50 
iterations, while the parameters are resampled every 100 iterations. The domain is 
sampled by selecting 20 spatial and 10 temporal points, forming a Cartesian product 
of 200 training locations. The PCE basis used in the PINN-PCE model consists of 
orthogonal polynomials up to degree 5. We observe that the training and testing errors 
of the PINN-PCE model are consistently higher than those of the plain PINN model. This 
behavior remains even when increasing the degree of the PCE basis.

\begin{figure}[tbhp]
\centering 
\subfloat[%
    Mean and standard deviation of the reference and PINN-based solutions.%
                ]{%
                \label{fig:diffusion:a}\includegraphics[width=0.6\textwidth]{%
                    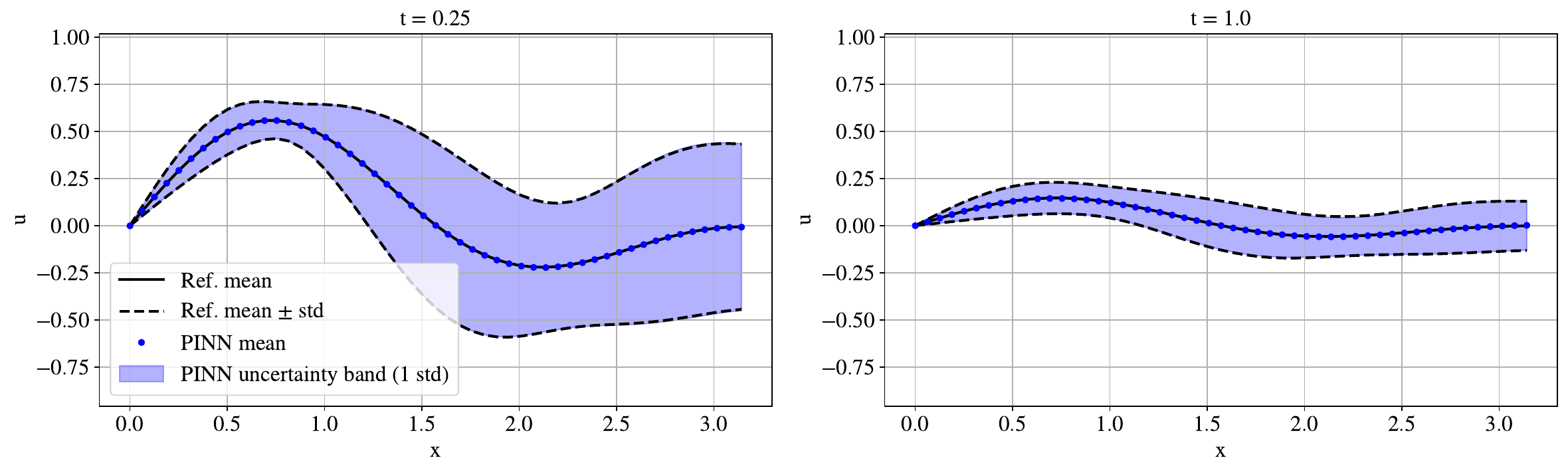}%
                }\\
\subfloat[%
    Average values comparison.%
                ]{%
                \label{fig:diffusion:b}\includegraphics[width=0.6\textwidth]{%
                    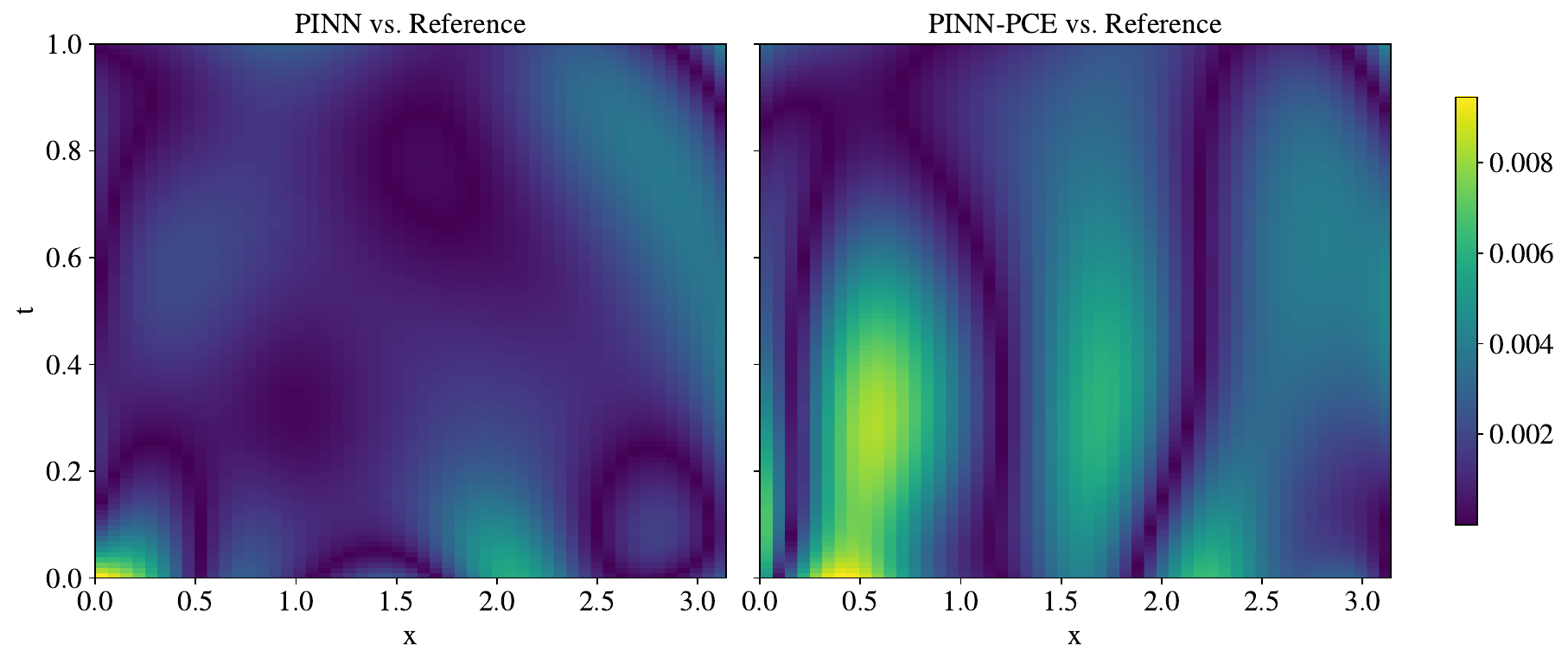}%
                }\\
\subfloat[%
    Histogram comparison at time $t=0.5$.%
                ]{%
                \label{fig:diffusion:c}\includegraphics[width=0.7\textwidth]{%
                    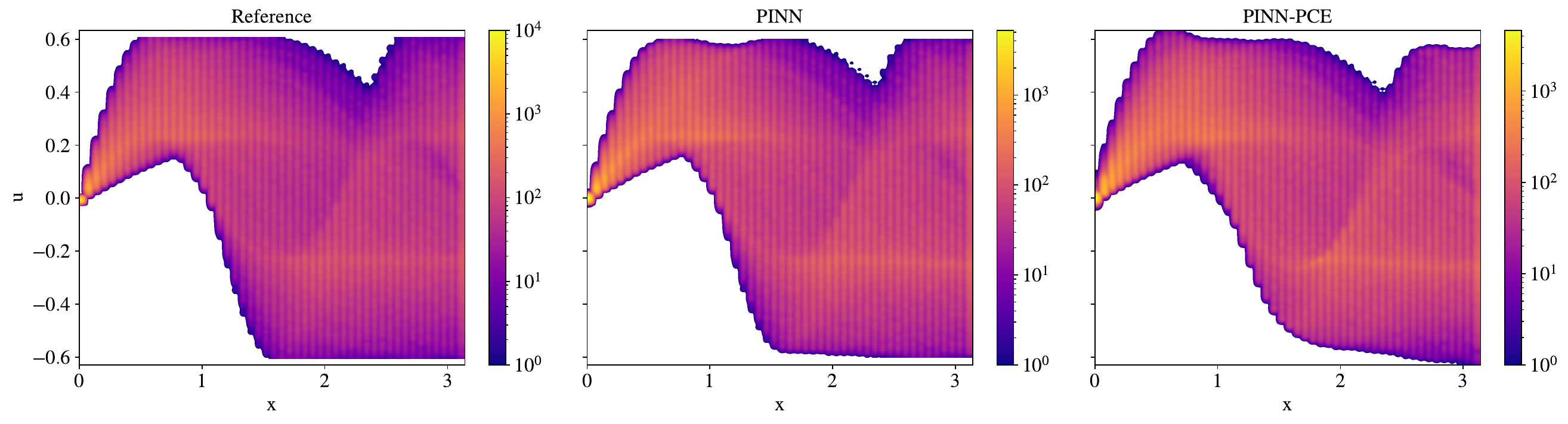}%
                }
\caption{%
    Average and one standard deviation of the reference and PINN-based solutions (a), 
    at time $t=0.5$ and $t=1$. Average values (b) and histograms of the solution (c) of 
    the diffusion equation~\cref{eq:diffusion} at time $t=0.5$ using the PINN and the 
    PINN-PCE architectures. 
    Notice that PINN produces marginally better results than the PINN-PCE architecture.
}
\label{fig:diffusion}
\end{figure}

In \Cref{fig:diffusion}, we present the average solution and histograms for the
diffusion equation \cref{eq:diffusion} at time $t = 0.5$, computed using the PINN and
PINN-PCE architectures. In \Cref{fig:diffusion:a}, the x-axis represents the spatial
coordinate and the y-axis corresponds to time. \Cref{fig:diffusion:b} displays the
spatial coordinate on the x-axis and the logarithm of the frequency of solution values on
the y-axis. We observe that the PINN-PCE architecture yields slightly improved results
compared to the plain PINN. Both methods show good qualitative agreement with the 
analytical solution, indicating their capacity to learn the underlying dynamics of the 
equation.

\subsection{Reaction-Diffusion}

We consider the partial differential equation,
\begin{equation}
  u_t - D u_{xx} + g u^3
  = f, \quad t \in [0, 4],\ x \in [-1, 1] ,
\end{equation}
with initial and boundary conditions,
\begin{gather} \label{eq:reaction_diffusion}
  u(0, x) = 0.5 \cos^2(\pi x) , \\
  u(t, -1) = u(t, 1) = 0.5 .
\end{gather}
The reaction function $g$ is given by,
\begin{equation}
  g(x) = 0.2 + e^{r_1 x} \cos^2(r_2 x),
\end{equation}
with $r_1 \sim \mathcal{U}(0.5, 1)$ and $r_2 \sim \mathcal{U}(3, 4)$
and the forcing function is given by,
\begin{equation}
  f(x) = \exp\left( -\frac{(x - 0.25)^2}{2 k_1^2} \right) \sin^2(k_2 x),
\end{equation}
with $k_1 \sim \mathcal{U}(0.2, 0.8)$ and $k_2 \sim \mathcal{U}(1, 4)$.
The diffusion coefficient is fixed to $D=0.01$.
This equation has been studied in \cite{Psaros:2023}.
%%%%%%%%%%%%%%%%%%
%%%%%%%%%%%%%%%%%%

In \Cref{fig:reaction_diffusion_loss}, we plot the training 
and testing loss functions during the training of the PINN architecture for the reaction-diffusion 
equation \cref{eq:reaction_diffusion}. The spatial-temporal domain is resampled every 50 
iterations, while the parameters are resampled every 100 iterations. The domain is 
sampled by selecting 40 spatial and 20 temporal points, forming a Cartesian product 
of 800 training locations. 

\begin{figure}[tbhp]
\centering 
\subfloat[%
    Mean and standard deviation of the reference and PINN-based solutions.%
                ]{%
                \label{fig:reaction_diffusion:a}\includegraphics[width=0.6\textwidth]{%
                    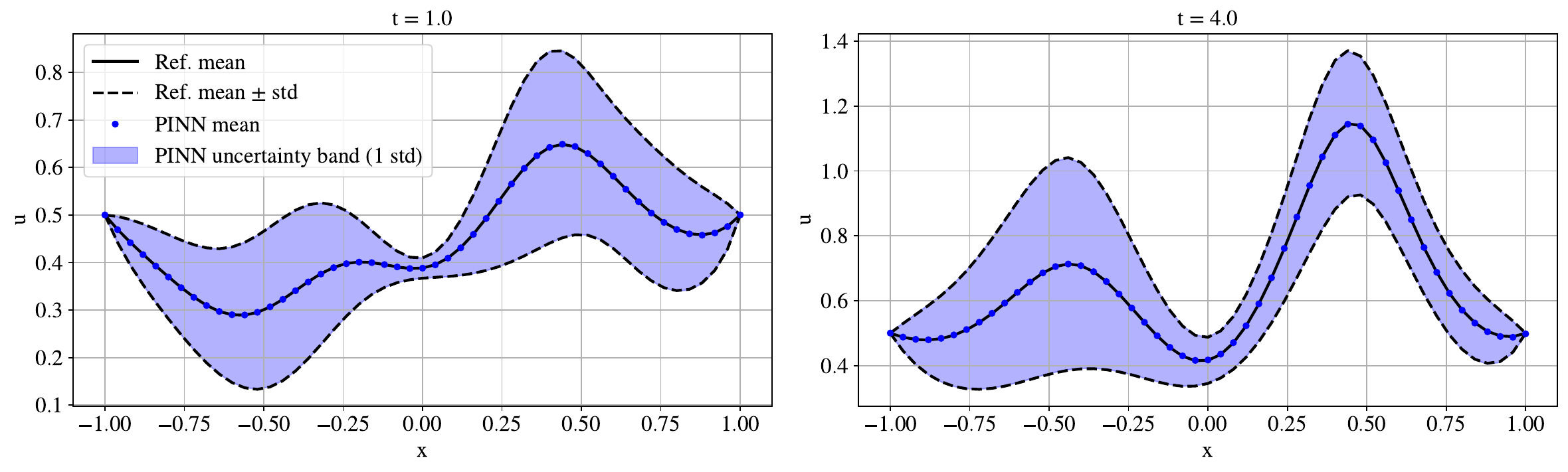}%
                }\\
\subfloat[%
    Absolute error heatmap.%
                ]{%
                \label{fig:reaction_diffusion:b}\includegraphics[width=0.6\textwidth]{%
                    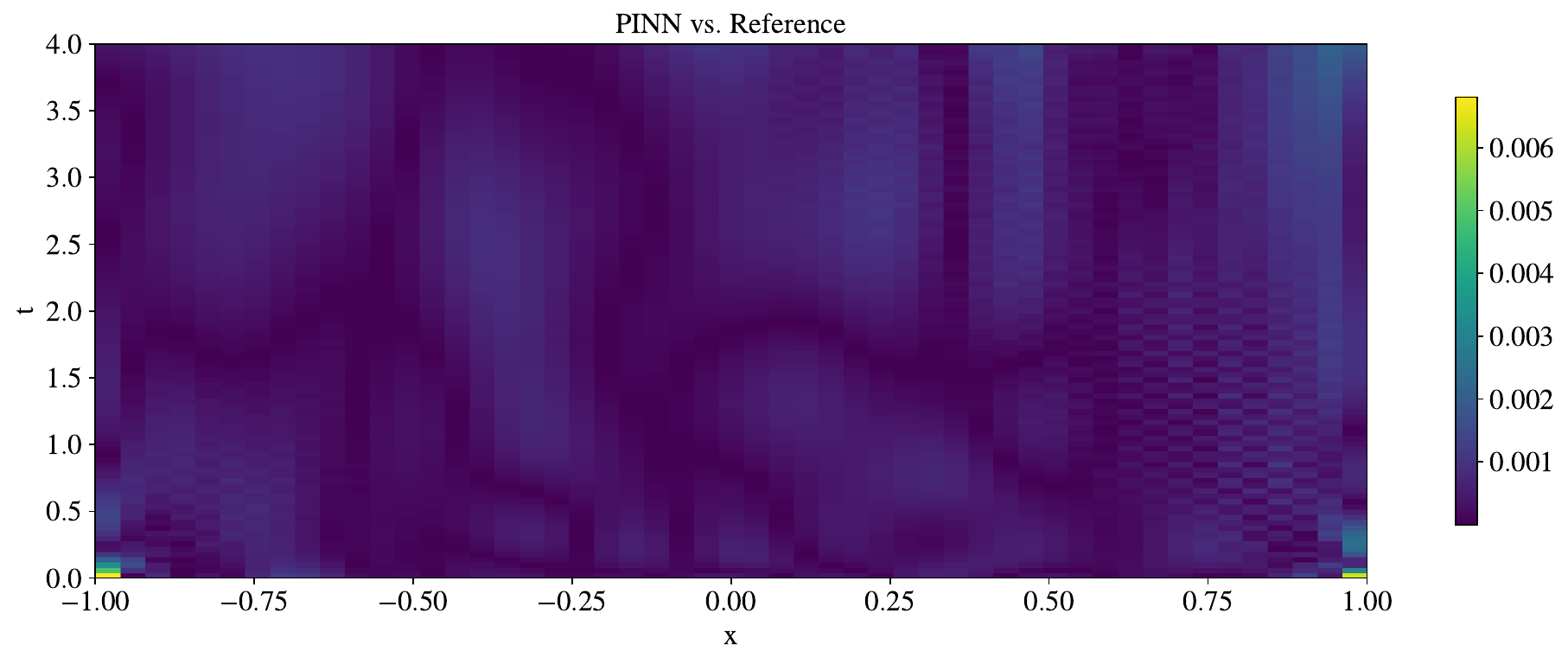}%
                }\\
\subfloat[%
    Histograms of the solution at time $t = 2$.%
                ]{%
                \label{fig:reaction_diffusion:c}\includegraphics[width=0.7\textwidth]{%
                    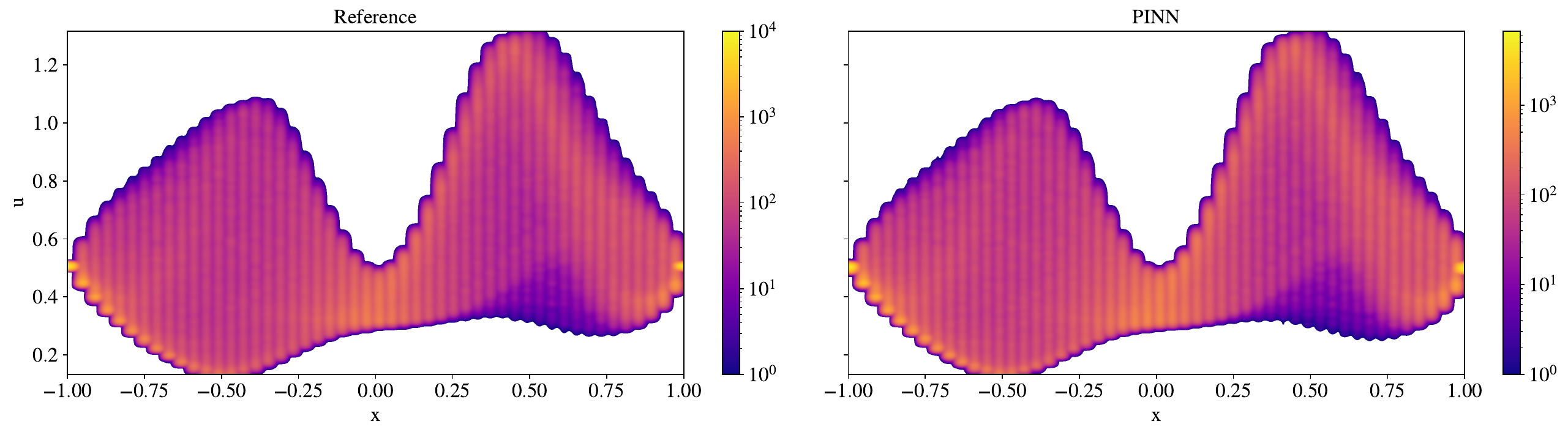}%
                }
\caption{%
    Average values and histograms of the solution of the reaction-diffusion 
    equation~\cref{eq:reaction_diffusion} at time $t = 2$ using the PINN architecture. 
    The statistics produced by the PINN architecture are in good qualitative agreement 
    with the reference solution.   
}
\label{fig:reaction_diffusion}
\end{figure}

In \Cref{fig:reaction_diffusion}, we plot the mean and standard deviation of the 
reference and PINN-based solutions. The absolute error heatmap shows that the PINN-based 
solution is in good qualitative agreement with the reference solution. The histograms of 
the solution at $t=2$ show that the PINN-based solution is in good qualitative agreement 
with the reference solution. The error shown in \Cref{fig:reaction_diffusion:b} shows
larger values at the boundaries of the domain for initial times. This flaw can probably be
fixed by using adaptive selection of the training points based on the error.

\appendix

\section{\!Universal Approximation Theorem}\label{Se:6}

\!\!Universal Approximation Theorems (UATs) are foundational results in the theory of 
neural networks. They establish that certain classes of 
functions—often defined by specific neural network architectures—are dense in a 
target function space. In other words, these theorems guarantee that the model 
class is expressive enough to approximate any function in the target space to an 
arbitrary degree of accuracy.

The classical Universal Approximation Theorem states that neural networks with 
a continuous, non-polynomial activation function are dense in the space of 
continuous functions on compact subsets $\R^d$ 
\cite{cybenko1989approximation,
    hornik1989multilayer,
    pinkus1999approximation}. 
Subsequent developments extended these results to approximation in Sobolev 
spaces and provided quantitative approximation rates for specific activations such 
as $\text{ReLU},$ $\text{ReLU}^k$ and $\tanh$ 
\cite{yarotsky2017error,
    petersen2018optimal,
    de2021approximation,
    xu2020finite}. 
However, in many applications -- and in particular for parametric PDEs -- the relevant 
function domains are non-compact, and the target functions take values in 
infinite-dimensional Banach spaces such as Sobolev spaces. In this direction, 
Neufeld and Schmocker \cite{neufeld2025universalapproximationresultsneural}
proved a universal approximation theorem on non-compact domains for networks 
with general non-polynomial activations, providing density results in appropriate  
weighted function spaces. Meanwhile, universal approximation results for 
operator-learning networks mapping between Banach-valued function spaces 
have been established in more specialised settings, such as Theorem 9.3 in 
Kovachki et al. \cite{Stuart_NOper_JMLR:24}, which applies to certain classes of 
neural operators.

Building on these developments, we show below that the neural-network-based spaces 
of probability measures we consider are capable of approximating probability measures 
represented through Bochner spaces $L^2(\gamma;\HH)$, where $\HH$ is a general 
Hilbert space (eg., $H^s(D)$). Thus, our Neural Measures spaces are expressive enough 
to yield accurate approximations for our purpose.

\subsubsection*{Revisiting UAT for Banach Spaces $X$}
Let $U$ a Borel subset of $\R^m$, and denote by $C_b^k(U;\R^d)$ the vector space of 
$k-$times continuously differentiable functions $f:U\to\R^d$ such that both the function 
and all its partial derivatives up to order $k$ are continuous and bounded. In the case 
$k\geq1$, we additionally assume that $U$ is open. 

For a multi-index $a=\lr{a_1,\dots a_m}$ with $|a|\leq k$, the $a-$th partial derivative is 
given by
\begin{equation}\label{derivative:a}
     u\mapsto \partial_a f(u) \coloneqq 
     \frac{\partial^{|a|}}{\partial u_1^{a_1}\dots\partial u_m^{a_m}} f(u) \in \R^d.
\end{equation}
Equipped with the norm
\begin{equation*}
    \norm{f}{C_b^k(U;\R^d)}\coloneqq  
    \max_{|a|\leq k} \, \sup_{u \in U} \abs{\partial_a f(u)},
\end{equation*}
the space $C_b^k(U;\R^d)$ becomes a Banach space.

To introduce weighted variants, we define the polynomially weighted $C^k-$norm:
\begin{equation*}
    \norm{f}{C_{pol,\beta}^k(U;\R^d)}\coloneqq  \max_{|a|\leq k} \, \sup_{u \in U} 
    \frac{\abs{\partial_a f(u)}}{\lr{1+\abs{u}}^\beta},
    \qquad  \text{ for some }\beta\geq 0.
\end{equation*}
Using this norm, we define the weighted $C^k-$space 
$\overline{C_b^k(U;\R^d)}^\beta$ as the closure of $C_b^k(U;\R^d)$ under the 
$\norm{\cdot}{C_{pol,\beta}^k(U;\R^d)}-$norm.
In either case (bounded, or unbounded $U$), this space is naturally identified as subset 
of $C_b^k(U;\R^d)$; for details, see Lemma 4.1 of 
\cite{neufeld2025universalapproximationresultsneural}. 
Thus, equipped with the weighted norm, 
$\lr{\overline{C_b^k(U;\R^d)}^\beta,\norm{\cdot}{C_{pol,\beta}^k(U;\R^d)} }$, is Banach 
space.
\begin{assumption}[for UAT in Banach Spaces]\label{Ass:UAT:Banach}
    Let $k\geq 0,\, U\subset \R^m,$ and $\beta>0$. 
    In the special case where $k\geq 1$, we assume $U$ is open.
    Let $\lr{X,\norm{\cdot}{X}}$ be a Banach space of functions $f:U\to\R^m$ such that 
    the restriction map
    \begin{equation*}
        \lr{ C_b^k(\R^m;\R^d),\norm{\cdot}{C_{pol,\beta}^k}}\ni f \mapsto f\Big|_U \in 
        \lr{X,\norm{\cdot}{X}},
    \end{equation*}
    is a continuous dense embedding\footnote{%
        \begin{minipage}[t]{0.95\linewidth}
            \!“Continuous dense embedding” \!means \!the \!restriction \!map 
            $\!i_U\!\!:\!\! \lr{\! C_b^k(\!\R^m;\!\R^d\!),\!\norm{\cdot}{C_{pol,\beta}^k}\!\!} \!\! \to \!\!
            \lr{\!X,\!\norm{\cdot}{X}\!}$ \\
            is a bounded linear injection with dense image, i.e.,  $\|i_U\|<\infty$ and 
            $\operatorname{cl}_{\norm{\cdot}{X}} \lr{Im(i_U)} = X$.
        \end{minipage}%
        }.
\end{assumption}
\begin{theorem}[Restating Theorem 2.7 in \cite{neufeld2025universalapproximationresultsneural}, 
    UAT for approximating Banach space $X$] \label{th:2.7:Restate}
    Suppose we are given parameters $k,\,\beta$, a Borel subset $U\subset\R^m$ and 
    a Banach space $X$ consisting of functions with domain $U$ 
    such that \Cref{Ass:UAT:Banach} holds.
    Let $\rho\in \overline{C_b^k(\R)}^\beta$ be a non-polynomial activation function, 
    and consider the following class of neural networks:
        \begin{equation}\label{NeuralNet}
           \NN_{\rho,\R^m,d}^{(K)} = \lrr{%
            \begin{split}
                  \phi:\R^m\to\R^d \text{ with } \phi(u) = 
                  \sum_{n=1}^K y_n \rho(a_n^T u - b_n),& \\
                  &~\hspace{-6cm}\text{ for some }
                  a_1,\dots a_K\in\R^m,\, b_1,\dots b_K\in\R,\,
                  y_1,\dots y_K\in\R^d
            \end{split}%
          },        
        \end{equation}
    and let $\NN_{\rho,\R^m,d}\coloneqq \bigcup_{K\in\N}\NN_{\rho,\R^m,d}^{(K)}$. 
    Then, it holds that 
    \begin{equation*}
        cl_{\norm{\cdot}{X}} \lr{\NN_{\rho,\R^m,d}\Big|_U}= X,
    \end{equation*}
    i.e., the closure of the neural network class $\NN_{\rho,\R^m,d}$ 
    (restricted to $U$) with respect to the $X-$norm coincides with $X$ itself.
\end{theorem}
Several examples of Banach spaces satisfying \Cref{Ass:UAT:Banach} are 
provided in Example 2.6 of \cite{neufeld2025universalapproximationresultsneural}.

\subsubsection*{UAT for PCE-NN/ Galerkin-NN Based Neural Measure Spaces}

In both versions of the Universal Approximation Theorem (UAT) for Neural Measure 
Space architectures introduced in \Cref{sec:truncated:basis,sec:GalerkinNN}, 
we work within the Bochner space $L^2(\gamma;\HH)$. In each 
case, the approximation results rely on expansions of this space: either in a basis of 
$L^2(\gamma)$ or in a basis of $\HH$. More generally, such expansions may also be 
formulated directly in a basis of $L^2(\gamma;\HH)$ itself.

A sufficient condition ensuring the existence of these bases is the separability of 
$L^2(\gamma;\HH)$. This is provided by the following assumption.
\begin{assumption}[Separability Conditions for $L^2(\gamma;\HH)$]\label{Ass:Spaces}
    We assume that $\gamma$ is a Borel probability measure defined on 
    $\Bor{\Xi}$ of a Polish space $\Xi$ and $\HH$ is a separable Hilbert 
    space.
\end{assumption}
Under \Cref{Ass:Spaces}, the separability of  $L^2(\gamma;\HH)$ follows 
from a sequence of classical results. Since $\gamma$ is a Borel probability measure on 
the Polish space $\Xi$, the space $L^2(\gamma)$ is a separable Hilbert space (see 
Proposition 3.4.5 in \cite{cohn2013measure}). 
By assumption, $\HH$ is also separable Hilbert space. Therefore, both $\HH$ and 
$L^2(\gamma)$ admit countable orthonormal bases, as guaranteed by 
Theorem 9 in \cite{lax2014functional}. 
Moreover, the algebraic tensor product $L^2(\gamma)\otimes \HH$ is dense in the 
Bochner space $L^2(\gamma;\HH)$ ( Lemma 1.2.19 of \cite{tuomas3analysis}). 
Combining these points yields the separability of $L^2(\gamma;\HH)$.

The algebraic tensor product used above is defined by
\[
    f\otimes g(\xi,x) = f(\xi)g(x),\quad \text{ for }f\in L^2(\gamma),\,g\in\HH,
\]
and finite linear combinations of such elements form $L^2(\gamma)\otimes \HH$. The 
inner product of $L^2(\gamma;H)$  respects the tensor structure:
\begin{equation}\label{eq:innerPro}
  \inner{f \otimes g}{\overline{f} \otimes \overline{g}}_{L^2(\gamma; \HH)} = 
  \inner{f}{\overline{f}}_{L^2(\gamma)} \inner{g}{\overline{g}}_{\HH},
  \quad \text{for all } f, \overline{f} \in L^2(\gamma), \ g, \overline{g} \in \HH.
\end{equation}

Since $L^2(\gamma;H)$ is separable, it admits a countable orthonormal basis 
$\lrr{\tilde{\phi}_n}_{n\in\N}\in L^2(\gamma;H)$. 
It is easy to verify that any $q\in L^2(\gamma;\HH)$  
can be represented either as
\begin{equation}\label{L2:repres:onH}
  q(x;\xi) = \sum_{n\in\N} a_{n}(\xi)\, \psi_n(x),\quad
          \text{ for some }\lrr{a_n}_{n\in\N} \subset L^2(\gamma),
\end{equation}  
or
\begin{equation}\label{L2:repres}
  q(x;\xi) = \sum_{n\in\N} b_{n}(x)\, \phi_n(\xi),\quad
          \text{ for some }\lrr{b_n}_{n\in\N} \subset H.
\end{equation}  
where $\lrr{\psi_n}_{n\in\N}$ and $\lrr{\phi_n}_{n\in\N}$ are orthonormal bases of $\HH$ 
and $L^2(\gamma)$, respectively. Both representations are understood in the 
$L^2(\gamma;\HH)$ sense.

\begin{theorem}[\!Universal Approximation via PCE-NN in Neural Measure Spaces]\label{Th:PCE:UAT}
    Suppose the Hilbert space $\HH$ satisfies \Cref{Ass:UAT:Banach} 
    for some $k,\,\beta,\,U\subset \R^{in}$, and consider 
    representation of $L^2(\gamma;\HH)$ given in \cref{L2:repres}. 
    Let $\rho\in \overline{C_b^k(\R)}^\beta$ to be a non polynomial activation function 
    and define the class of neural networks $\NN_{\rho,\R^{d_{\text{in}}},d_{\text{out}}}^{(K)}$ as in 
    \cref{NeuralNet}. For integers $K,\, M\geq 1$, set 
    \begin{equation*}
          \VV_\NN^{(K,M)} \coloneqq \lrr{%
          \begin{split}
           	g_\theta: \R^{d_{\text{in}}}\to\R^{d_{\text{out}}M}: \,\text{ for some }
           	g_\theta =\lr{g_\theta^{(1)},\dots,g_\theta^{(M)}}& \\
          	&~\hspace{-3cm} \text{ with } 
           	g_\theta^{(1)},\dots,g_\theta^{(M)}\in \NN_{\rho,\R^{d_{\text{in}}},d_{\text{out}}}^{(K)}
          \end{split}
          },
    \end{equation*} 
    and define the associated class of Neural Measures by 
    \begin{equation*}
          \VV_\NN^{(K,M),*} = \lrr{%
          \begin{split}
          \pushforward{X_\theta}{\gamma}\in\PPP_2^\gamma\lr{\,\,\HH\,\,}: \text{ where }
          X_\theta(\xi)=\sum_{n=1}^{M } g_\theta^{(n)}\,&\phi_n(\xi)\,\,\\
          &\text{ for some }g_\theta\in \VV_\NN^{(K,M)}
          \end{split}
          }.%
    \end{equation*}
    Finally, let 
    \begin{equation*}
          \hat{\VV}_\NN^* = \bigcup_{K,M \in \N} \VV_\NN^{(K,M),*} .
    \end{equation*}
    Then, the closure of $\hat{\VV}_\NN^*$ with respect to the $W_2-$metric on 
    $\PPP_2^\gamma(\HH)$ coincides with the entire space $\PPP_2^\gamma(\HH)$.
    In other words, for every $\nu \in  \PPP_2^\gamma(\HH)$, there exists a sequence
    $\lrr{\mu_m}_{m\in\N} \subset \hat{\VV}_\NN^*$ such that 
    \begin{equation*}
          \lim_{m\to\infty} W_2(\nu,\mu_m) = 0.
    \end{equation*}
\end{theorem}
\begin{remark}
    \!In \Cref{Th:PCE:UAT}, parameters $(\!K,\!M)$ act as hyper-parameters. 
    Moreover, recall 
    that in \Cref{sec:truncated:basis} we fixed a specific neural network 
    architecture $\VV_\NN$, 
    i.e.,  we fixed $K,\,M$ in $\VV_\NN^{(K,M)}$. 
    Thus, the class $\VV_\NN^*$ defined in \cref{PCE:VNstar} corresponds exactly to 
    the specialised case $\hat{\VV}_\NN^{(K,M),*}$ for some fixed $K,\,M$, under the 
    notation of this section.
\end{remark}
\begin{proof}
    For any $\mu,\,\nu\in \PPP_2^\gamma(\HH)$, by definition  
    there exist two functions $g,\,f\in L^2\lr{\gamma;\HH}$ such that
    \begin{equation*}
    \mu(B) = \pushforward{g}{\gamma}(B),\quad\nu(B) = \pushforward{f}{\gamma}(B),
    \qquad \forall B\in\Bor{\HH}.
    \end{equation*}
    Considering the 2-Wasserstein distance $W_2\lr{\nu,\mu}$ on $\HH$
    (with the norm-induced metric), we have
    \begin{equation}\label{W_2:Prop:Upp:Bound}
     \begin{split}
     	W_2\lr{\nu,\mu}^2  &= 
	\inf_{\pi\in\Pi\lr{\nu,\mu}}\int_{\HH\times\HH} \norm{h_1-h_2}{\HH}^2\,
	\pi\lr{\ud h_1,\ud h_2}\\ 
     	&\leq \int_{\Xi} \norm{f(\xi) - g(\xi) }{\HH}^2\,\gamma\lr{\ud \xi}  = 
	\norm{f- g}{L^2\lr{\gamma;\HH}}^2.
     \end{split}
    \end{equation}
    The inequality holds since the joint distribution  
    $\pushforward{\lr{f,g}}{\gamma}$ is contained 
    within plans $\Pi\lr{\nu,\mu}$.
    
    Now fix $\nu\in \PPP_2^\gamma(\HH)$, let $f\in L^2\lr{\gamma;\HH}$ be such 
    that $\nu = \pushforward{f}{\gamma}$. Using the expansion \cref{L2:repres}, define
    truncated approximations
    \begin{equation}\label{rewrite:Repres}
        f^{(m)}(x;\xi)=\sum_{n\leq n_m} \tilde{b}_{n}(x) \phi_n\lr{\xi},
    \end{equation}
    where $\lrr{\phi_n}_{n\in\N}$ is an orthonormal bases of $L^2(\gamma)$, 
     $\lrr{\tilde{b}_n}_{n\in \N}\subset \R$, and $n_m$ is an increasing 
    sequence  such that
    \begin{equation*}
        \norm{f-\sum_{n\leq n_m} \tilde{b}_{n}\phi_n }{
        L^2(\gamma;\HH)} \leq  \frac{1}{2\, m}.
    \end{equation*}

    By \Cref{Ass:UAT:Banach,th:2.7:Restate}, we get the
    closedness of neural networks,  i.e., 
    \[
        \operatorname{cl}_{\norm{\cdot}{\HH}} 
        \lr{\NN_{\rho,\R^{d_{\text{in}}},d_{\text{out}}}\Big|_U}= \HH.
    \]
    Thus, we can approximate each coefficient function $\tilde{b}_n$ with neural network 
    $g_{n}^{(m)}$ so that  
    \begin{equation}\label{Choice:GN}
        \max_{n\leq n_m} \norm{\tilde{b}_{n} - g_n^{(m)}}{\HH} <\frac{1}{2\,n_m\,m},
    \end{equation}
    and define 
    \[
        X_m(x;\xi) =\sum_{n\leq n_m} g_n^{(m)}\,\phi_n(\xi),\quad 
                        \mu_m \coloneqq \pushforward{X_m}{\gamma}.
    \] 
    
    Then, the $\lrr{\mu_m}_{m\in\N}$ sequence satisfies 
    \begin{equation}\label{control:W:with:L2}
    \begin{split}
         W_2\lr{\nu,\mu_m} &\leq \norm{f- X_m}{L^2\lr{\gamma;\HH}} \\
         &\leq 
         \norm{f- f^{(m)}}{L^2\lr{\gamma;\HH}}  + \norm{f^{(m)}- X_m}{L^2\lr{\gamma;\HH}} 
         \leq \frac{1}{m},
    \end{split}
    \end{equation}
    where in the last inequality, we use \cref{Choice:GN} and the triangle inequality.
\end{proof}

In the case of Galerkin-NN setting, we shall verify below the density on 
$\PPP_2^\gamma(\HH_h)$, where $\HH_h$ is a Galerkin type subspace of 
$\HH$. The density on $\PPP_2^\gamma(\HH)$ will depend on the approximation 
Properties of $\HH_h$ and once these are given it will follow by similar arguments 
as in the proof of \Cref{Th:PCE:UAT}. 

To this end,  $\HH_h\subset \HH$ be a fixed finite dimensional Galerkin subspace 
of dimension $M$, with basis $\lrr{\psi_n}_{n=1}^M$. Suppose a probability space 
$\lr{\R^{d_{\text{ran}}},\Bor{\R^{d_{\text{ran}}}},\gamma}$ is given such that 
$L^2(\gamma)$ satisfies \Cref{Ass:UAT:Banach} for some 
hyper-parameters\footnote{Example 2.6 of \cite{neufeld2025universalapproximationresultsneural}
specifies suitable hyperparameters and properties of $\gamma$.
}
$k,\,\beta$, and let $U\coloneqq supp\lr{\gamma}\subset \R^{d_{\text{ran}}}$. Under the representation 
\cref{L2:repres:onH}, the space $L^2(\gamma;\HH_h)$ reduces to the finite expansion 
\begin{equation}\label{Galerkin:base:2}
  q(x;\xi) = \sum_{n=1}^M a_n(\xi) \psi_n(x) ,
\end{equation}
where $\psi_n$ are not necessarily orthonormal.

Under these conditions, we consider Neural Measure Spaces as follows:
Let $\rho\in \overline{C_b^k(\R)}^\beta$ be a non-polynomial activation function.
We define class of Neural Measures with finite $K\geq 1$ as
\begin{equation*}
      \VV_\NN^{(K),*} = \lrr{\begin{split}
          \pushforward{X_\theta}{\gamma}\in\PPP_2^\gamma\lr{\,\,\HH_h\,\,}: 
          \text{ where }
          X_\theta(\xi)=\sum_{n=1}^{M} g_\theta^{(n)}(\xi)\,\psi_n&, \\
      &~\hspace{-3cm}\text{for some }
      g_\theta^{(1)},\dots,g_\theta^{(M)}\in \NN_{\rho,\R^{d_{\text{ran}}},1}^{(K)}
      \end{split}
      }.
\end{equation*}
based on the Neural Networks $\NN_{\rho,\R^{d_{\text{ran}}},1}^{(M)}$ defined in 
\cref{NeuralNet}.

Let us now verify that 
\[
    \hat{\VV}_\NN^* = \bigcup_{K \in \N} \VV_\NN^{(K),*}
\]
is dense in $\PPP_2^\gamma(\HH_h)$ with respect to the $W_2-$metric.

For fixed $\nu\in \PPP_2^\gamma(\HH_h)$, let $f\in L^2\lr{\gamma;\HH_h}$ be such 
that $\nu = \pushforward{f}{\gamma}$. According to expansion \cref{Galerkin:base:2}, 
we define the coefficient function $\tilde{a}_{n}$ of $f$ by
\begin{equation*}
    f(x;\xi)=\sum_{n=1}^M \tilde{a}_{n}(\xi) \psi_n\lr{x}.
\end{equation*}

Since $L^2(\gamma)$ satisfies \Cref{Ass:UAT:Banach,th:2.7:Restate} 
provides the following closedness result:
\[
    \operatorname{cl}_{\norm{\cdot}{L^2(\gamma)}} \lr{\NN_{\rho,\R^{d_{\text{ran}}},1}\Big|_U}= L^2(\gamma). 
\]
Therefore, there exists a sequence of $\lrr{g_{1}^{(m)},\dots g_{M}^{(m)}}_{m=1}^\infty$ with $g_{n}^{(m)}\in \NN_{\rho,\R^{d_{\text{ran}}},1}$ such that 
\begin{equation}
    \max_{n\leq M} \norm{\tilde{a}_{n} - g_n^{(m)}}{L^2(\gamma)} <
    \frac{1}{m \sum_{n=1}^M \|\psi_n\|_{\HH}}.
\end{equation}
Based on the sequence of $g_{n}^{(m)}$'s, we define 
\[
    X_m(x;\xi) =\sum_{n=1}^M g_n^{(m)}(\xi)\,\psi_n(x),\quad \text{ and set }\quad 
                    \mu_m \coloneqq \pushforward{X_m}{\gamma}.
\] 
By applying Inequality \cref{W_2:Prop:Upp:Bound}, we have
\begin{equation*}
     W_2\lr{\nu,\mu_m}\leq \norm{f- X_m}{L^2\lr{\gamma;\HH}} \leq 
     \max_{n\leq M} \norm{\tilde{a}_{n} - g_n^{(m)}}{L^2(\gamma)}\,\,
            \sum_{n=1}^M \|\psi_n\|_{\HH}< \frac{1}{m}.
\end{equation*}
confirming the convergence $\mu_m\to \nu$ with respect to $W_2-$metric.

\subsubsection*{UAT for Fully Network-Based Neural Measure Spaces}

In the setting of Fully Network-Based Neural Measure Spaces, we now consider a 
Hilbert space $\HH$ tied to 
our original problem. Specifically, we take $\HH=W^{k,2}(D;\R^{d_{\text{out}}})$, 
the Sobolov space of $\R^{d_{\text{out}}}-$valued functions on a bounded open domain 
$D\subset\R^{d_{\text{in}}}$, defined as:
\begin{equation*}
    W^{k,2}(D;\R^{d_{\text{out}}})\coloneqq \lrr{f \in L^2\lr{D; \R^{d_{\text{out}}}}: \,\,\norm{f}{k,2}^2 = 
    \sum_{|a|\leq k} \norm{\partial_a f}{L^2\lr{D; \R^{d_{\text{out}}}}}^2<\infty  },
\end{equation*}
where $ \partial_a f$ denotes the weak derivative of $f$ on some multi-index $a$ 
(as defined in equation \cref{derivative:a}). 
Since $L^2\lr{D; \R^{d_{\text{out}}}}$ is a separable Hilbert space, it follows that 
$W^{k,2}(D;\R^{d_{\text{out}}})$ is also separable.

Let $\gamma$ be a Borel probability measure on $\Bor{\Xi}$ with 
$\Xi\subset \R^{d_{\text{ran}}}$. 
Define the probability measure $\lambda$ on $\R^{d_{\text{in}}}$, as 
$\lambda(\ud x) = \frac{1}{\abs{D}}1(x\in D)\ud x$ 
and consider the product measure $\gamma\otimes\lambda$ on $\Xi\times D$.

Now define the function space
\begin{equation} 
    \Lambda^{(0,k),2}(\gamma\otimes\lambda; \R^{d_{\text{out}}}) = 
    \lrr{%
    \begin{split}
        f \in L^2\lr{\gamma\otimes\lambda; \R^{d_{\text{out}}}}&: \\
        &~\hspace{-3.3cm}\norm{f}{(0,k),2}^2 = \lr{
            \sum_{|a|\leq k} \norm{\partial_a f}{L^2\lr{\gamma\otimes\lambda; \R^{d_{\text{out}}}}}^2
        }^{1/2}<\infty,\\
        &~\hspace{-3.3cm}\text{ with multi-indices } a =\lr{a_1,\dots a_{d_{\text{ran}}}, a_1',\dots a_{d_{\text{in}}}' }\\
        &~\hspace{-2cm}\text{ and } a_i = 0\text{ for every }i\leq d_{\text{ran}}
    \end{split}
    }
\end{equation}
i.e., we only differentiate with respect to the input variables $x\in D$.

By applying Fubini’s theorem on functions $f$ which either in 
$L^2\!\!\lr{\!\gamma;\!W^{k,2}\!\lr{\!D;\R^{d_{\text{out}}\!}}}$, 
or in $\Lambda^{(0,k),2}(\gamma\otimes\lambda; \R^{d_{\text{out}}})$, we obtain 
\begin{equation*}
    \norm{f}{L^2(\gamma;W^{k,2}(D;\R^{d_{\text{out}}}))}  = 
    \abs{D} \norm{f}{\Lambda^{(0,k),2}(\gamma\otimes\lambda; \R^{d_{\text{out}}})}.
\end{equation*}
From which, we get that these two spaces are  isometrically isomorphic, in our 
case these spaces are identical.

Under an appropriate choice of the hyperparameters 
$k,\,\beta,\,\Xi\times D\subset \R^{d_{\text{in}}+d_{\text{ran}}},\,\gamma$, 
one can ensure that $\Lambda^{(0,k),2}(\gamma\otimes\lambda; \R^{d_{\text{out}}})$ satisfies 
\Cref{Ass:UAT:Banach}. 
Suitable choices of these hyperparameters can be obtained by combining the 
conditions listed in Example 2.6 of 
\cite{neufeld2025universalapproximationresultsneural}.

Now let $\rho\in \overline{C_b^k(\R)}^\beta$ to be a non-polynomial activation function,
and define the class of neural networks $\NN_{\rho,\R^{d_{\text{in}}+d_{\text{ran}}},d_{\text{out}}}^{(M)}$ 
as in \cref{NeuralNet}. By \Cref{th:2.7:Restate}, we then obtain the 
approximation result:
\begin{equation*}
    \operatorname{cl}{\norm{\cdot}{X}} \lr{\NN_{\rho,\R^{d_{\text{in}}+d_{\text{ran}}},d_{\text{out}}}\Big|_{\Xi\times D}}= 
    L^2(\gamma;W^{k,2}(D;\R^{d_{\text{out}}})),
\end{equation*}
where norm $\norm{\cdot}{X}$ stands for $L^2(\gamma;W^{k,2}(D;\R^{d_{\text{out}}}))-$norm.

Finally, similar to the proof of \Cref{Th:PCE:UAT} -- 
and more precisely, by applying inequality \cref{control:W:with:L2} which shows that 
the $L^2-$norm controls the $W_2-$metric 
-- we obtain the Universal Approximation Theorem for Fully Network-Based Neural 
Measure Spaces when the target Hilbert space is a Sobolev space. That is, for every 
$\nu \in  \PPP_2^\gamma(\HH)$, there exists a sequence 
$\lrr{\mu_m}_{m\in\N} \subset \hat{\VV}_\NN^*$ such that 
\begin{equation*}
      \lim_{m\to\infty} W_2(\nu,\mu_m) =0.
\end{equation*}
%

%%%%%%Supplementary Materials Start%
\subsection{\texorpdfstring{%
Well-Definedness of $U(\xi)$ Distribution%
}{Well-Definedness of U(xi) Distribution}%
}
\label{well-def}

Based on the Random PDE stated in \Cref{eq:RPDE:v2}, we are interested in approximating
its probability distribution, namely $\pushforward{U}{\gamma}$. We would like to know
under what assumptions can we ensure that this distribution is well-defined. Moreover,
does $\pushforward{U}{\gamma}$ depend on the choice of the underlying probability space
$\lr{\Omega,\FF,\p}$?

To answer this, we first define two mappings. Given a stochastic process
\begin{equation*}
  f:D\times\Xi\to \R\quad \text{ with }\quad f(\cdot;\xi)\in \tilde\HH,
\end{equation*}
and
\begin{equation*}
  g:\Gamma\times\Xi\to \R\quad \text{ with }\quad g(\cdot;\xi)\in \overline\HH,
\end{equation*}
we define the evaluation map
\begin{equation*}
  Q_{f,g}:\Xi\to \tilde\HH\times\overline\HH\times\Xi\quad \text{ with }\quad \xi\mapsto
  \lr{f(\cdot;\xi),g(\cdot;\xi),\xi}.
\end{equation*}
Next, we introduce the solution operator
\begin{equation*}
  G: \tilde\HH\times\overline\HH\times\Xi \to \HH \text{ with } \lr{\tilde{f},\tilde{g},
  \tilde\xi}\mapsto \tilde{u}(\cdot;\tilde\xi) ,
\end{equation*}
where $\tilde{u}(\cdot;\xi)$ is the solution to the Random PDE (similar to the one in
\Cref{eq:RPDE:v2}) for a fixed $\xi =\tilde\xi$, and with right hand side replaced by
$\tilde{f}$ and $\tilde{g}$.

Then, the solution $U(\xi)$ to the original problem can be represented as:
\begin{equation*}
  U(\xi) = G(f(\cdot;\xi),g(\cdot;\xi),\xi) = G \circ Q_{f,g}(\xi).
\end{equation*}
A sufficient condition for ensuring both the well-definedness of $\pushforward{U}{\gamma}$
and its invariance with respect to the abstract probability space $\lr{\Omega,\FF,\p}$ is
the following:
\begin{itemize}
\item The solution operator $G$ is well-defined for every point in
$\tilde\HH\times\overline\HH\times \Xi$, and
\item The composition $G\circ Q_{f,g}(\xi)$ is measurable on each instance of
$\lr{f,g}\in L^2(\gamma;\tilde\HH)\times L^2(\gamma;\overline\HH)$.
\end{itemize}
Under these conditions, it follows that for any two probability spaces
$\lr{\Omega,\FF,\p}$ and $\lr{\tilde\Omega,\tilde\FF,\tilde\p}$, with corresponding
random elements $\xi:\Omega\to\Xi$ and $\tilde\xi:\tilde\Omega\to\Xi$, both having
law $\gamma$, we have:
\begin{equation*}
  \pushforward{U\circ\xi}{\p}=\pushforward{G\circ Q_{f,g}\circ\xi}{\p}=\pushforward{G\circ
  Q_{f,g}}{\gamma}=\pushforward{G\circ Q_{f,g}\circ\tilde\xi}{\tilde\p} = \pushforward{U\circ
  \tilde\xi}{\p}.
\end{equation*}
This confirms that the push-forward measure $\pushforward{U}{\gamma}$ is independent of the
choice of the underlying probability space, provided the stated conditions are met.

\subsection{Discrete Neural Network Spaces}

We describe our notation regarding the generic discrete neural network spaces which we
denote by $V _{\mathcal{N}}.$ These are spaces approximating functions
\begin{equation*}
  v : \R ^ {d_{I}} \to \R ^{d_{O}} .
\end{equation*}
We have used different choices for $d_{I}$ and $d_{O}$ throughout this
paper. To fix ideas, we present a generic structure of the spaces $V _{\mathcal{N}}.$ Of
course several alternative architectures are possible.
A \emph{deep neural network} maps every point
$\overline y\in \R ^ {d_{I}}$ to $v_\theta (\overline y) \in \R ^ {d_{O}}$, through
\begin{equation}\label{C_L}
	v_\theta(\overline y)= \mathcal{C}_L (\overline y) :=C_L \circ \sigma \circ C_{L-1} \cdots
	\circ\sigma \circ C_{1} (\overline y) \quad \forall \overline y \in \R ^ {d_{I}}.
\end{equation}

The intermediate  layers $C_k$,   are affine maps of the form
\begin{equation}
	C_k y =
  W_k y +b_k, \qquad \text{where }  W_k \in \R ^ {d_{k+1}\times d_k}, b_k \in \R ^ {d_{k+1}},
\end{equation}
where the dimensions $d_k$ may vary with each layer $k$ and $\sigma (y)$ denotes the
vector with the same number of components as $y$, where
$\sigma (y)_i= \sigma(y_i).$ The index $\theta$ represents collectively all
the parameters of the network. The space of parameters is denoted by $\Theta$:
\begin{equation*}
  \Theta=\lrr{\theta:=\lr{b_1,W_1,\dots W_L, b_L}: \text{ for }b_i\in\R^{d_k},\,W_i\in
  \R^{d_{k+1}\times d_k}, \ k=1, \dots, L }.
\end{equation*}
The set of networks $\mathcal{C}_L$ with a given architecture of the form \cref{C_L},
\cref{C_k} is called $\mathcal{N}$.
Then $\Theta \in \R ^ {\dim{\mathcal {N}}}$ where
the total number of degrees of freedom of $\mathcal {N} ,$ is $\dim{\mathcal {N}}= \sum
_{k=1} ^L d_{k+1} (d_k +1)  .$ The nonlinear discrete set of functions $V _{\mathcal{N}}$
is defined as
\begin{equation}
	V _{\mathcal{N}}= \{ u_\theta : \varOmega _T \to \R ,  \ \text{where }  u_\theta (\x)
  = \mathcal{C}_L (\x), \ \text{for some } \mathcal{C}_L\in  \mathcal{N}\, \}  .
\end{equation}

\subsection{Neural Networks with Random Input as Generative Models}

In generative modeling, neural networks receive both a standard input and a random
input (often called the latent code or noise). Formally, assume a probability distribution
$\gamma$ over $\R^{d_{\text{ran}}}$ (also known as the seed or reference distribution),
and a neural network $g_\theta\in\VV_\NN$. Then, the generative model is the random
variable:
\begin{equation*}
  g_\theta(x;\xi), \qquad\text{ where } \xi\sim \gamma.
\end{equation*}
This induces a probability distribution over outputs, defined via the pushforward
measure:
\begin{equation*}
  g_\theta(x;\xi) \sim \pushforward{ g_\theta(x,\cdot)}{\gamma} .
\end{equation*}
To simplify notation, define:
\begin{equation*}
  Q_{\theta,x}(\xi):= g_\theta(x,\xi)
\end{equation*}
Then, the output distribution for a fixed input $x$ is:
\begin{equation*}
  \mu_\theta(x,\,A) :=
  \gamma\circ Q_{\theta,x}^{-1} (A) =
  \int_{\mathbb{R}^{d_{\text{ran}}}}
  \mathds{1}_{\{g_\theta(x,\xi)\in A\}} \,\gamma(\ud \xi),
  \qquad\forall A\in\Bor{\R^{d_{\text{out}}}}.
\end{equation*}
If the mapping $g_\theta: \R^{d_{\text{in}}}\times\R^{d_{\text{ran}}}\to\R^{d_{\text{out}}}$ is
measurable with respect to the Borel $\sigma-$algebras
$\Bor{\R^{d_{\text{in}}}\times\R^{d_{\text{ran}}}}$ and $\Bor{\R^{d_{\text{out}}}}$, then the
following properties hold:
\begin{itemize}
\item
For every fixed $x\in\R^{d_{\text{in}}}$, $\mu_\theta(x,\cdot)$ is a probability measure
on $\R^{d_{\text{out}}}$
\item
For every fixed $A\in\Bor{\R^{d_{\text{out}}}}$, the function
$x\mapsto \mu_\theta(x,\,A)$ is measurable with respect to $\Bor{\R^{d_{\text{in}}}}$.
\item
$\mu_\theta(\cdot, A)$ is $\Bor{\R^{d_{\text{in}}}}/\Bor{\R^{d_{\text{out}}}}-$measurable
\end{itemize}
Therefore, $g_\theta$ defines a stochastic kernel $\mu_\theta(x,\,\ud y)$ from
$\R^{d_{\text{in}}}$ to $\R^{d_{\text{out}}}$.

To establish these properties, one can utilise the fact that
\begin{equation*}
  \Bor{\R^{d_{\text{in}}}\times\R^{d_{\text{ran}}}}= \Bor{\R^{d_{\text{in}}}}\otimes\Bor{\R^{d_{\text{ran}}}},
\end{equation*}
since both $\R^{d_{\text{in}}}$ and $\R^{d_{\text{ran}}}$ are Polish spaces, see  Theorem 17.25
in \cite{kechris2012classical}.

\begin{figure}[h]
  \centering
  \includegraphics[width=0.7\textwidth]{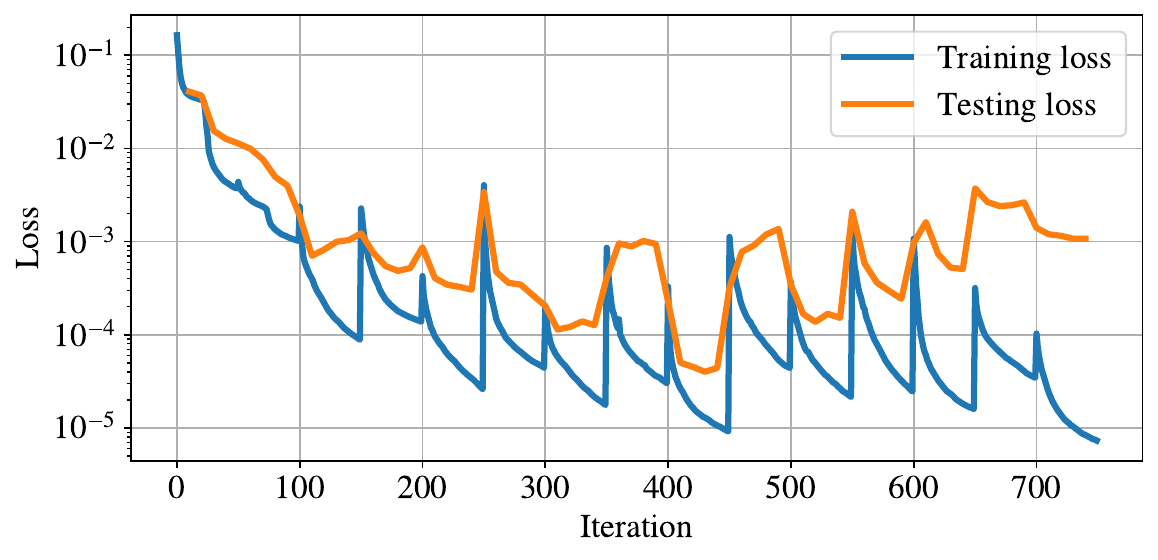}
  \caption{Training and testing loss over time during the training of the PINN
  architecture for the bistable ODE~\cref{eq:bistable_ode}.}
  \label{fig:bistable_ode_loss}
\end{figure}

\begin{figure}[h]
  \centering
  \includegraphics[width=0.7\textwidth]{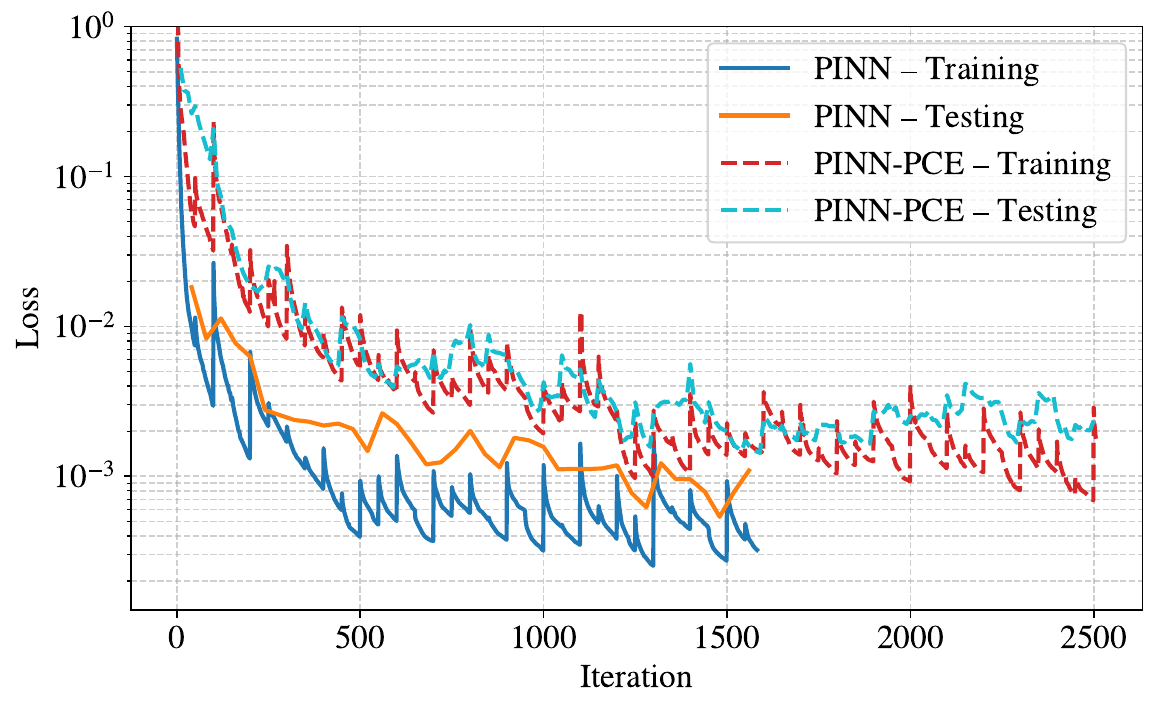}
  \caption{Training and testing loss function over time of the training of the PINN and
  the PINN-PCE architecture for the diffusion equation \cref{eq:diffusion}.}
  \label{fig:diffusion_loss}
\end{figure}

\begin{figure}[t]
  \centering
  \includegraphics[width=0.7\textwidth]{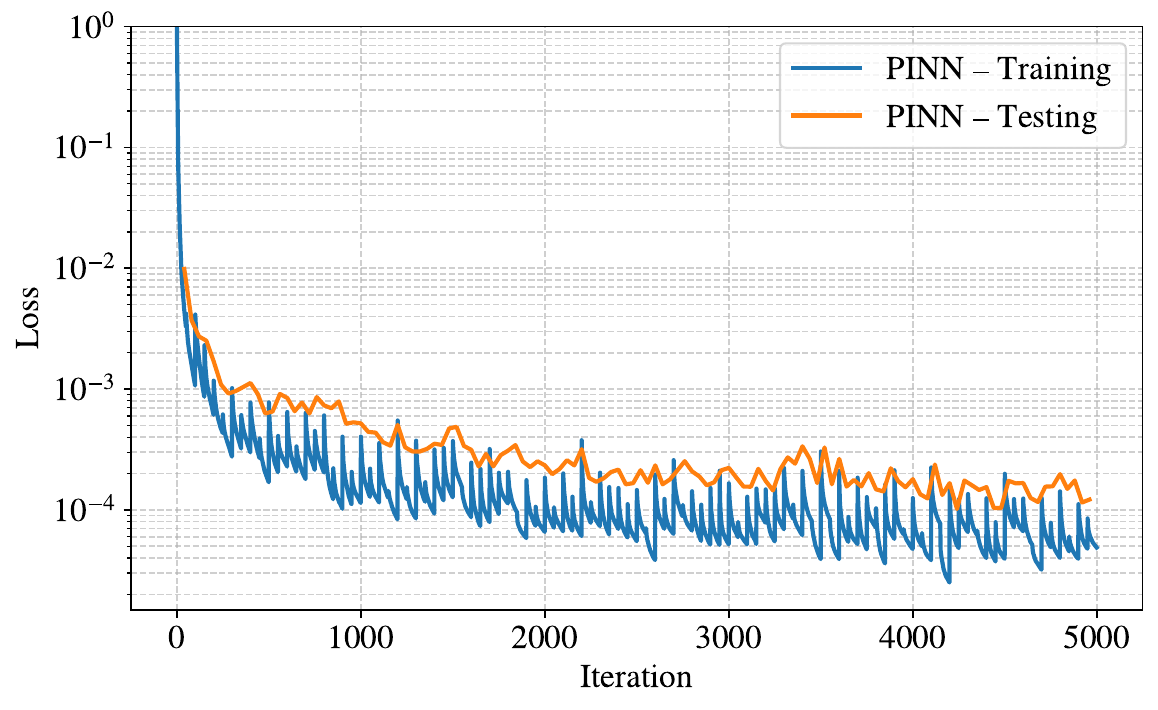}
  \caption{Training and testing loss function over time of the training of the PINN and
  the PINN-PCE architecture for the reaction-diffusion equation \cref{eq:reaction_diffusion}.}
  \label{fig:reaction_diffusion_loss}
\end{figure}

%%%%%%Supplementary Materials End %%%%

\section*{Acknowledgments}

The authors used AI-assisted tools (ChatGPT by OpenAI) for minor editorial 
assistance, including grammar correction and language polishing. All substantive 
ideas, analysis, and results presented in this work are entirely the authors’ own.

\bibliographystyle{siamplain}
\bibliography{bibliography}
\end{document}

%% file: shared.tex
\usepackage{amsfonts,amsmath, amssymb, amsopn}
\usepackage{mathtools}
\usepackage{dsfont}
\usepackage{bm, mathrsfs}
\usepackage[only,llbracket,rrbracket]{stmaryrd}
\usepackage{cleveref}   
\usepackage{xcolor}
\usepackage{graphicx}
\usepackage[caption=false]{subfig}

\DeclareMathOperator*{\argmin}{arg\,min}
\newcommand{\lr}[1]{\left( #1 \right)}
\newcommand{\lrr}[1]{\left\{ #1 \right\}}
\newcommand{\lrbr}[1]{\left[ #1 \right]}
\newcommand{\Bor}[1]{\mathcal{B}\lr{#1}}
\newcommand{\ud}{{\rm d}}
\newcommand{\p}{\mathbb{P}}
\newcommand{\FF}{\mathcal{F}}
\newcommand{\HH}{\mathcal{H}}
\newcommand{\PPP}{\mathcal{P}}

\newcommand{\VV}{\mathcal{V}}

\newcommand{\NN}{\mathcal{N}}
\newcommand{\X}{\mathscr{X}}
\newcommand{\x}{\bm{x}}
\newcommand{\R}{\mathbb{R}}
\newcommand{\N}{\mathbb{N}}
\newcommand{\E}{\mathbb{E}}

\newcommand{\pushforward}[2]{\lrbr{#1}_{\!\#} #2}
\newcommand{\norm}[2]{\left\| #1 \right\|_{#2}}
\newcommand{\abs}[1]{\left|#1\right|}
\newsiamremark{remark}{Remark}
\newsiamremark{assumption}{Assumption}

\headers{Neural Measures for Learning Distributions of Random PDEs}{G. Arampatzis, S. Katsarakis, and C. Makridakis}
\title{Neural Measures for Learning Distributions of Random PDEs}
\author{Georgios Arampatzis\footnotemark[2]\,\,\,\thanks{DMAM, University of Crete, Greece.}
\and Stylianos Katsarakis\thanks{IACM, Foundation for Research and Technology - Hellas, Greece.}
\and Charalambos Makridakis\footnotemark[2]\,\,\,\thanks{MPS, University of Sussex, United Kingdom.	}
}

\ifpdf
\hypersetup{
  pdftitle={Neural Measures for Learning Distributions of Random PDEs},
  pdfauthor={G. Arampatzis, S. Katsarakis, and C. Makridakis}
}
\fi

\newcommand{\exteq}[2][]{%
  {\color{siaminlinkcolor}\eqref{#2}%
  \ifx\\#1\\%
  \else\ #1\fi}%
}
\newcommand{\extfig}[2][]{%
  {\color{siaminlinkcolor}Figure \ref{#2}%
  \ifx\\#1\\%
  \else\ #1\fi%
  }%
}

\DeclarePairedDelimiterX{\inner}[2]{\langle}{\rangle}{#1, #2}